\documentclass[11pt]{article}

\usepackage[final]{acl}

\usepackage{times}
\usepackage{latexsym}
\usepackage{amsmath}
\usepackage{amssymb}

\usepackage[T1]{fontenc}
\usepackage[utf8]{inputenc}

\usepackage{microtype}
\usepackage{inconsolata}
\usepackage{graphicx}
\usepackage{booktabs}
\usepackage{array}
\usepackage{multirow}
\usepackage{tabularx}
\usepackage{soul}

\usepackage[most]{tcolorbox}

\newtcolorbox{promptbox}[1][]{%
  enhanced, breakable, colback=black!3, colframe=black!45, boxrule=0.4pt,
  arc=2pt, left=5pt, right=5pt, top=4pt, bottom=4pt,
  fonttitle=\bfseries\footnotesize, coltitle=black, colbacktitle=black!10,
  attach boxed title to top left={xshift=6pt, yshift=-2pt},
  boxed title style={colframe=black!45, boxrule=0.4pt, arc=1pt},
  #1}

\definecolor{hlyellow}{RGB}{255,241,153}
\sethlcolor{hlyellow}

\title{Evaluating Persona Consistency in LLM User Simulators Without External Reference Annotations}

\author{
  Anping Zhang\textsuperscript{1,2} \quad
  Yang Tan\textsuperscript{1} \quad
  Yipeng Liu\textsuperscript{1,3} \quad
  Clive Bai\textsuperscript{1} \quad
  Blaze Chen\textsuperscript{1} \quad
  Yang Li\textsuperscript{3} \\[0.3em]
  \textsuperscript{1}Tencent \quad
  \textsuperscript{2}Tsinghua University \quad
  \textsuperscript{3}The Chinese University of Hong Kong, Shenzhen \\}

\begin{document}
\maketitle
\begin{abstract}
Standard evaluations assess LLM-simulated personas against an external ground truth, such as community trait annotations, but that reference is unavailable when the persona is inferred only from a short, task-oriented chat. We ask how persona consistency can be evaluated without such a reference, and propose a self-anchored \emph{internal-consistency diagnostic}: the persona describes itself once and acts twice. A Likert self-report supplies a soft internal anchor; two behavioral profiles come from constrained scenario choices scored without a judge and from open-ended responses judged by an LLM. Consistency is the correlation between the self-report and each behavioral profile, in a shared six-facet BFI-2 space. The framework compares persona-construction strategies by assessing how consistently their elicited responses reflect their self-reported traits. Across ordinary-user chats (WildChat and LMSYS) and well-documented personas with community annotations (the Personality Database, PDB), profile-enriched representations agree with the anchor more closely than a raw-dialogue baseline. On PDB, this self-anchored ordering reproduces the community-annotation ordering. The pattern is stable across answering and judge models.
\end{abstract}

\begin{figure}[t]
    \centering
    \includegraphics[width=\linewidth]{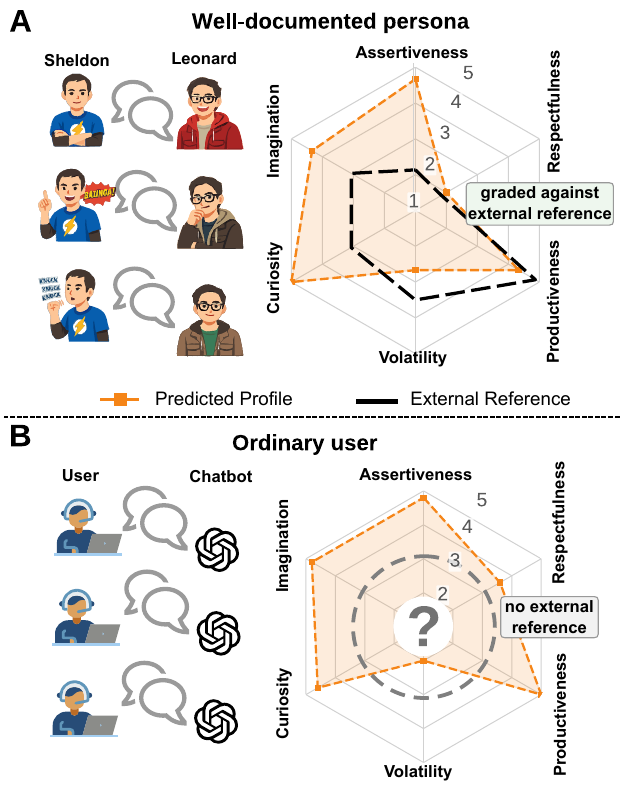}
    \caption{Evaluating a simulated persona: a well-documented persona versus an ordinary user. (A) For a well-documented persona (e.g., Sheldon Cooper), the predicted six-facet profile can be graded against an external reference. (B) For an anonymized, conversation-derived ordinary user (WildChat/LMSYS), no external reference exists, so the predicted profile cannot be graded in the same way, motivating our self-anchored framework.}
    \label{fig:intro-gt-gap}
\end{figure}

\section{Introduction}
A \emph{user simulator} is a large language model (LLM) agent instructed to act as a particular user or persona. Such simulators are becoming core infrastructure for building and studying LLM assistants~\citep{chen2024persona,wang-etal-2025-know}: they provide interactive counterparts for training and evaluating assistants~\citep{dou-etal-2025-simulatorarena}, synthesize dialogue data at scale~\citep{wang-etal-2025-know}, and stand in for human participants in field-style studies of user preference and behavior~\citep{li-etal-2025-far,wang2024incharacter}. Their usefulness depends on whether the simulator expresses persona-relevant preferences and behaviors consistently. This is more feasible to assess for well-documented public personas, whose traits can be reconstructed from abundant public material (encyclopedia entries, biographies, recorded interviews and speeches). Ordinary users create an \emph{information asymmetry}: their personality cannot be reliably reconstructed from a short, task-dominated chat history~\citep{zhao2024wildchat,zheng2024lmsys}, so conversation-derived personas leave no external personality reference to grade a simulation against. Evaluating persona consistency for these personas therefore remains an open problem, and it is the setting this paper studies.

Existing role-play and persona evaluations typically compare simulated behavior with an explicit persona specification, an external character reference, or matched human behavior: the dominant \emph{role-play fidelity} paradigm grades a simulated character against an annotated script, persona profile, or psychometric-interview reference~\citep{wang2024incharacter}, while simulator-evaluation methods instead compare against matched human dialogue or behavior sequences~\citep{dou-etal-2025-simulatorarena,li-etal-2025-far}. None of these is generally available for personas constructed from sparse, task-oriented ordinary-user conversations. Self-/temporal-consistency methods avoid this requirement but only certify a stable \emph{surface} profile (name, occupation, stated facts), not the deeper trait-level preferences that actually govern how a user reacts; the remaining evaluation approaches target orthogonal properties (fluency, helpfulness, factuality) and do not assess persona consistency at all. Without an evaluation signal, we also cannot compare persona representations—such as the raw conversation and an extracted profile—to identify which yields more consistent trait expression. Consequently, persona representations cannot be systematically evaluated or improved in this setting.

We therefore focus on a practically important question: how to evaluate persona consistency without an external reference. A simulator may reproduce persona traits in self-description without expressing them in situational responses. We treat agreement between the two as an \emph{internal-consistency diagnostic}: a simulator that has internalized its persona should let the traits it reports govern how it acts, whereas one that merely restates its prompt should not. We implement this idea as a \emph{self-anchored evaluation framework}. A Likert self-report over a chat-relevant six-facet subset of the BFI-2~\citep{soto2017bfi2}---scored at the facet rather than the domain level, because facets carry behavioral information that domain averaging discards~\citep{paunonen2001behavior}---provides a \emph{soft internal anchor} elicited from the persona itself. The same persona then acts in two structurally different channels: a deterministically scored scenario questionnaire (SCN) and LLM-judged open-ended responses (OE). All three profiles occupy the same six-facet space. Consistency is the \emph{correlation between each behavioral profile and the self-report anchor}, per persona across facets or per facet across personas. Because all three profiles come from the same model, we first check that their agreement is not only a shared model or prompt effect. We then apply the framework to PDB, WildChat and LMSYS---the latter two with no external user-level reference---and test whether it distinguishes construction strategies. The metric compares those strategies; it does not establish human fidelity or downstream utility.

Our main contributions are summarized as follows:
\begin{itemize}
    \item We frame persona-consistency evaluation without external user-level labels as an internal-consistency problem for conversation-derived user simulation, and propose a \emph{self-anchored, cross-method diagnostic} over a chat-relevant six-facet BFI-2 subset (\S\ref{sec:method}).
    \item We show that the diagnostic separates persona representation strategies on all three datasets, stably across answering and judge models, and that on PDB the self-anchored ordering tracks the community-annotation ordering \mbox{(\S\ref{sec:results})}.
\end{itemize}

\section{Related Work}

\paragraph{Persona and role-play evaluation.} InCharacter grades role-playing agents via psychological interviews checked against an external character reference~\citep{wang2024incharacter}; PersonaGym probes persona adherence through dynamically generated persona-relevant environments~\citep{samuel-etal-2025-personagym}; Eval4Sim compares generated conversations to a human dialogue dataset along adherence, consistency, and naturalness~\citep{bao2026eval4sim}. Each requires an explicit persona specification, an external reference, or a matched human dataset, none of which is available for conversation-derived user personas built from sparse dialogue. We instead measure how strongly elicited behavior correlates with an internally elicited self-report anchor.

\paragraph{User simulation and its evaluation.} LLM user simulators are increasingly conditioned on profiles derived from interaction history, e.g., USP's implicit profiles~\citep{wang-etal-2025-know}, two of whose representation strategies we adopt (\S\ref{sec:construction}). SimulatorArena and BehaviorChain instead evaluate simulators against matched human messages or behavior sequences~\citep{dou-etal-2025-simulatorarena,li-etal-2025-far}. We address the intermediate problem of whether a conversation-derived persona is internally coherent when no such matched human reference exists.

\paragraph{Self-report validity and facet-level measurement.} Self-reported LLM traits should not automatically be treated as a behavioral reference: stable self-reports do not reliably predict behavior~\citep{han2025personality}, coherence is construct- and context-dependent~\citep{kocielnik2026rethinking}, and self-reports can share variance with LLM judges that human ratings do not~\citep{contreras2026llmnative}. This motivates our soft self-report anchor alongside two differently structured, behavior-oriented channels (\S\ref{sec:method}), evaluated through their convergent/discriminant structure rather than treated as a reference. We further operationalize personality at the facet level, since facets carry behavioral information that domain averaging discards~\citep{paunonen2001behavior,soto2017bfi2,deyoung2007aspects}, and since personality-inferability from public chat datasets is strongly interaction-dependent~\citep{zhao2024wildchat,zheng2024lmsys}, motivating our chat-relevant six-facet subset (\S\ref{sec:facets}).

\section{The Self-Anchored Persona Consistency Evaluation Framework}
\label{sec:method}
\begin{figure*}[t]
    \centering
    \includegraphics[width=\textwidth]{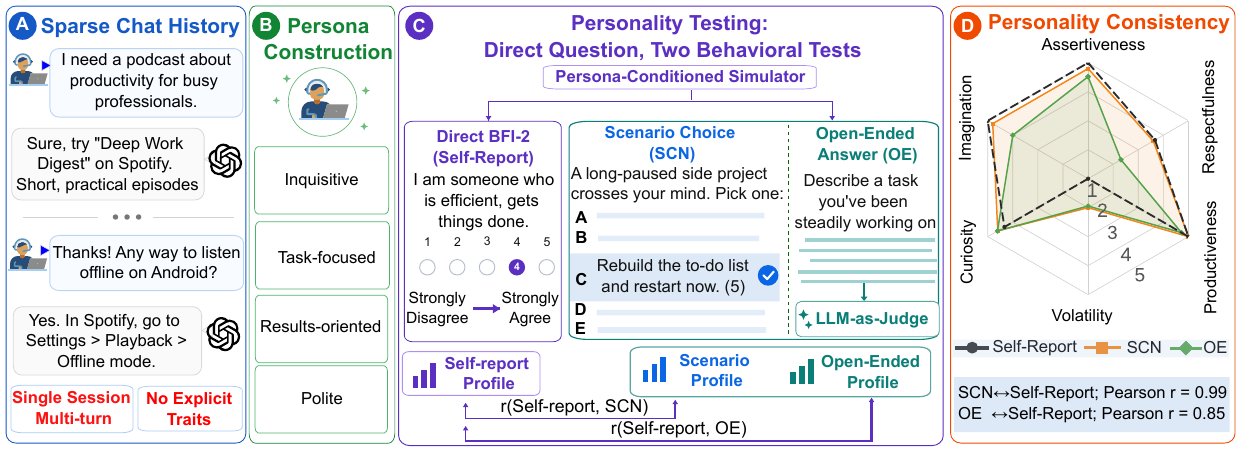}
    \caption{Overview of the self-anchored framework. A conversation-derived persona is evaluated through self-report, scenario choices (SCN), and open-ended responses (OE); consistency is measured as the correlation of the SCN and OE profiles with the self-report anchor. In (C), $r$ is the Pearson correlation between the persona's six-facet self-report profile and its SCN or OE profile.}
    \label{fig:framework}
\end{figure*}

The framework lets the persona describe itself once and act twice, then scores consistency as the \emph{correlation} between the self-report and the two behavioral channels (Figure~\mbox{\ref{fig:framework}}). We first fix a six-facet scoring space \mbox{(\S\ref{sec:facets})}, then elicit the self-report \mbox{(\S\ref{sec:step1})} and the two behavior-oriented profiles \mbox{(\S\ref{sec:step2})}, and finally compute the consistency statistic \mbox{(\S\ref{sec:step3})}. Because all three profiles come from the same model, we also validate the anchor \mbox{(\S\ref{sec:anchor-support}; Appendices~\ref{sec:appendix-mtmm}, \ref{sec:appendix-identifiability}, \ref{sec:appendix-persona-robust})}.

\subsection{Facet Space: Six Chat-Relevant Facets}
\label{sec:facets}
The Big Five Inventory--2 (BFI-2;~\citealp{soto2017bfi2}) operationalizes the Big Five personality model and organizes personality into five domains (Extraversion, Agreeableness, Conscientiousness, Negative Emotionality, and Open-Mindedness), each split into three facets (15 total; Appendix~\ref{sec:appendix-bfi2} lists them all). We retain six facets, one per domain plus a second Open-Mindedness facet: \emph{Assertiveness}, \emph{Respectfulness}, \emph{Productiveness}, \emph{Emotional Volatility}, \emph{Intellectual Curiosity} and \emph{Creative Imagination}. These six facets define the measurement space of the framework: every profile is scored on them, on a common $1$--$5$ scale.

We work at the \emph{facet} rather than the \emph{domain} level for incremental predictive validity over domain averaging~\citep{paunonen2001behavior,soto2017bfi2,deyoung2007aspects}. The subset is a design choice, not an empirically optimized selection. \emph{Channel observability} (primary): the facet should be elicitable in ordinary user--chatbot dialogue without contrived probing. Guided by topic distributions from public chat datasets \mbox{\citep{zhao2024wildchat,zheng2024lmsys}}, we drop facets that rarely arise there (e.g., Aesthetic Sensitivity, which typically needs talk of art or beauty) and keep facets expected to be more observable, such as Assertiveness, Productiveness, and Creative Imagination. \emph{Within-domain redundancy}: where sibling facets overlap in chat (e.g., Organization versus Productiveness), we keep one representative per domain. \emph{Expected identifiability}: conceptually distinct facets can still be hard to tell apart in text, so we streamline domains that would add little separable information, while retaining both Open-Mindedness facets because they cover complementary aspects of open-mindedness. Leave-one-facet-out checks leave the SCN Pearson strategy ranking against PDB unchanged (all Kendall $\tau=1$; Appendix~\ref{sec:appendix-lofo}). That supports ranking robustness \emph{within this selected space}; it does not establish psychometric reliability or validity, and coverage of personality remains limited (Limitations).

\subsection{Self-Anchored Report}
\label{sec:step1}
The first step produces the profile that everything else is compared against. The persona answers an 18-item Likert questionnaire derived from the BFI-2 (3 items $\times$ 6 retained facets), five times with independently shuffled item order; the facet score is the \emph{median} of the five administrations. Shuffling and the median limit item-order effects and single-run noise. This profile occupies the role an external reference would play in a standard evaluation, which is why we call it an anchor. \mbox{\S\ref{sec:anchor-support}} motivates that use. The instrument is documented in \mbox{Appendix~\ref{sec:appendix-likert}}.

\subsection{Two Behavioral Channels}
\label{sec:step2}
The second step asks the persona to act rather than describe itself, in two channels that probe the same six facets but fail differently. In \emph{scenario-based elicitation} (SCN), the persona chooses among options in 30 situational items (5 per facet); each option has a predefined score, so responses are scored by lookup with no LLM judge. In \emph{open-ended elicitation} (OE), the persona writes free-form answers to 18 prompts (3 per facet); an LLM judge scores each from $1$ to $5$ against a fixed rubric, in three independent calls that are then averaged. SCN uses constrained choices and deterministic scoring; OE uses free text and a judge. Agreement with the anchor in both channels is harder to dismiss as a format artifact, and the two methods enable the discriminant-validity analysis in \mbox{\S\ref{sec:anchor-support}}. Any format that returns one score per facet on the same scale could be substituted; the framework only needs at least two structurally dissimilar behavioral channels. Instruments, judge template, and rubric are in \mbox{Appendix~\ref{sec:appendix-scn}--\ref{sec:appendix-rubric}}.

\subsection{Evaluate Consistency}
\label{sec:step3}
Let \mbox{$a_i(f)$} and \mbox{$b_i(f)$} be persona \mbox{$i$}'s anchor and behavioral scores on facet \mbox{$f$}, with \mbox{$b\in\{\mathrm{SCN},\mathrm{OE}\}$}. Consistency is the Pearson correlation
\begin{equation}
r(x,y)=\frac{\sum_{k}(x_{k}-\bar{x})(y_{k}-\bar{y})}{\sqrt{\sum_{k}(x_{k}-\bar{x})^{2}\sum_{k}(y_{k}-\bar{y})^{2}}}
\label{eq:pearson}
\end{equation}
between the two, read in either of two directions. For a fixed facet we correlate over the personas of a dataset, giving a \emph{per-facet} correlation \mbox{$r^{b}_{f}$} that compares facets; for a fixed persona we correlate over its six facets, giving a \emph{per-persona} correlation \mbox{$r^{b}_{i}$} that compares personas, representations, or answering models \mbox{(Figure~\ref{fig:framework})}. On PDB, where each persona also carries an external annotation \mbox{$g_i(f)$}, we compute the same correlations against \mbox{$g$} in place of \mbox{$a$}, both for the two behavioral profiles and for the anchor itself. This quantifies the anchor's agreement with the external reference rather than assuming the anchor is accurate. Because Pearson $r$ is invariant to location and scale, we also report Lin's concordance correlation coefficient (CCC; \citealp{lin1989ccc}), which discounts $r$ for mean shift and scale mismatch, alongside Pearson in Tables~\ref{tab:main-results}--\ref{tab:facets-anchor}. We treat Pearson as the primary ranking statistic and use CCC to qualify absolute agreement. Correlations are aggregated over personas, channels, representations, and answering models in Fisher-\mbox{$z$} space,
\begin{equation}
\bar{r}=\tanh\!\Big(\tfrac{1}{|S|}\textstyle\sum_{s\in S}\operatorname{atanh}r_{s}\Big),
\label{eq:fisher}
\end{equation}
rather than by averaging \mbox{$r$} directly. The persona is always the sampling unit for uncertainty: analytic Fisher-\mbox{$z$} \mbox{$95\%$} confidence intervals (CIs) for a single across-persona correlation, and a by-persona cluster bootstrap when one persona contributes several non-independent rows \mbox{(Appendix~\ref{sec:appendix-persona-robust})}.

\subsection{Why the Self-Report Can Serve as an Anchor}
\label{sec:anchor-support}
The anchor is elicited from the same model as the two behavioral profiles, so a high correlation among them could reflect a shared response style rather than a persona-specific trait profile. We therefore test the anchor before relying on it. Existing checks support specificity but cannot rule out shared-model bias: we treat the metric as an internal-consistency diagnostic, not an absolute measure of fidelity.

On PDB, where community trait annotations exist, the anchor can be graded exactly as a behavioral profile would be: its six facet scores correlate with the external annotations at an average of \mbox{$r=0.62$} (where \mbox{$0$} would mean the anchor says nothing about the persona; \mbox{Table~\ref{tab:anchor-vs-ref}}), above the \mbox{$0.57$} and \mbox{$0.52$} that the scenario and open-ended profiles reach on the same personas. Personas that match their own anchor more closely also match the reference more closely \mbox{(\S\ref{sec:ref-results})}.

A further check asks whether the anchor separates the six facets. Following the \mbox{Campbell--Fiske} \mbox{multitrait--multimethod} criteria \mbox{\citep{campbellfiske1959}}, each facet agrees with itself across the three channels more than different facets do in \mbox{$68\%$}, \mbox{$75\%$} and \mbox{$63\%$} of tests on PDB, WildChat and LMSYS \mbox{(Appendix~\ref{sec:appendix-mtmm})}. A response style that moved all six facets together would not produce this separation. A second check confirms that it is also the \emph{persona} that is separated: a persona's behavior fits its own anchor better than any other persona's \mbox{(Appendix~\ref{sec:appendix-identifiability})}. Neither MTMM nor identifiability identifies which part of the agreement is persona-specific rather than shared-model style.

\begin{table}[t]
\centering\small
\setlength{\tabcolsep}{3pt}
\begin{tabular*}{\columnwidth}{@{\extracolsep{\fill}}lc lc@{}}
\toprule
\textbf{Facet} & \textbf{$r$} & \textbf{Facet} & \textbf{$r$} \\
\midrule
Assertiveness  & 0.49 & Emot.\ Volatility   & 0.70 \\
Respectfulness & 0.71 & Intell.\ Curiosity  & 0.58 \\
Productiveness & 0.66 & Creat.\ Imagination & 0.55 \\
\midrule
\textbf{Mean facet} & \multicolumn{3}{c}{\textbf{0.62}} \\
\bottomrule
\end{tabular*}
\caption{Per-facet Pearson \mbox{$r$} between the self-report anchor and PDB's external reference across the \mbox{$75$} personas.}
\label{tab:anchor-vs-ref}
\end{table}

\section{Experimental Setup}
\label{sec:setup}
\paragraph{Datasets.}
We use three datasets (Table~\ref{tab:datastats}). \emph{PDB}\footnote{Personality Database, \url{https://www.personality-database.com/}; community votes retrieved on 30 June 2026.} contains well-documented public personas ($40$ fictional characters and $35$ real public figures) with community-derived Big Five \emph{domain} scores, which we map to the six retained BFI-2 facets as domain-level proxies rather than facet-level ground truth (Appendix~\ref{sec:appendix-pdb}). We report the fictional and real subsets separately there. \emph{WildChat}~\citep{zhao2024wildchat} and \emph{LMSYS}~\citep{zheng2024lmsys} supply conversation-derived personas with no external user-level annotations; from each corpus we sample $100$ English conversations with at least eight user turns. There the analysis unit is a \emph{single conversation}, with no claim that a persona is a verified unique natural person; the USP representations~\citep{wang-etal-2025-know} are therefore \emph{per-conversation} rather than per-user longitudinal profiles.

\begin{table}[t]
\centering\small
\setlength{\tabcolsep}{5pt}
\begin{tabular}{l c cc}
\toprule
 & & \multicolumn{2}{c}{\textbf{Conversation-derived}} \\
\cmidrule(lr){3-4}
 & \textbf{PDB} & \textbf{WildChat} & \textbf{LMSYS} \\
\midrule
Analysis unit        & persona & \multicolumn{2}{c}{conversation} \\
\# units             & 75      & 100 & 100 \\
\addlinespace
Min.\ user turns     & ---     & \multicolumn{2}{c}{$\geq 8$} \\
External reference   & yes     & \multicolumn{2}{c}{no} \\
\# strategies        & 5       & \multicolumn{2}{c}{4} \\
\bottomrule
\end{tabular}
\caption{Dataset statistics.}
\label{tab:datastats}
\end{table}

\paragraph{Persona representation strategies.}
\label{sec:construction}
Table~\ref{tab:strategies} lists the strategies, which differ in where the persona information comes from and how explicitly it states traits. Two conversation-based strategies are available everywhere~\citep{chen2024persona,cheng-etal-2024-dialogues}; PDB supports three further strategies that anonymous personas do not permit, whereas WildChat and LMSYS instead use two representations from the USP profile-extraction pipeline~\citep{wang-etal-2025-know}. The three synthesized representations---Dialogue-driven Free Description, USP Attributes-only and USP Profile---are produced offline by a single extractor model, \texttt{GPT-5.5}~\citep{gpt55}, and not by the answering models, so every answering model receives the same representation text. Appendix~\ref{sec:appendix-construction} gives the full construction details.

\begin{table}[t]
\centering\small
\setlength{\tabcolsep}{3pt}
\renewcommand\tabularxcolumn[1]{m{#1}}
\begin{tabularx}{\columnwidth}{@{}>{\centering\arraybackslash}m{52pt} c >{\centering\arraybackslash}X@{}}
\toprule
\textbf{Source} & \textbf{Strategy} & \textbf{Input form} \\
\midrule
\multirow{4}{=}{\centering\emph{Conversation}}
  & Raw Dialogue        & Verbatim turns \\
  & Dialogue-driven     & Free text \\
  & USP Attributes-only & Structured fields \\
  & USP Profile         & Fields $+$ free text \\
\addlinespace
\multirow{2}{=}{\centering\emph{Persona material}}
  & Name-only           & Name \\
  & Wiki-based RAG      & Encyclopedic text \\
\addlinespace
\emph{Annotation}
  & Trait-injected      & Name $+$ traits \\
\bottomrule
\end{tabularx}
\caption{Persona representation strategies, grouped by where the persona information comes from~\mbox{\citep{wang-etal-2024-rolellm,wang-etal-2025-know}}. Traits are implicit in the first two groups and stated directly only under Trait-injected, which is therefore an upper reference point. Raw Dialogue and Dialogue-driven are used on all three datasets, the two USP strategies on WildChat and LMSYS, and the remaining three on PDB. On PDB, names are redacted except under Name-only and Trait-injected; WildChat and LMSYS personas are anonymized throughout.}
\label{tab:strategies}
\end{table}

\paragraph{Answering and judge models.}
We evaluate four answering models: \texttt{GLM-5.1-FP8}~\citep{glm51fp8}, \texttt{Hy3-preview}~\citep{hy3preview}, \texttt{gemma-4-31B-it}~\citep{gemma431bit}, and \texttt{gpt-oss-120b}~\citep{gptoss120b}. \texttt{DeepSeek-V4-Pro}~\citep{deepseekv4pro} is the default OE judge; we re-score the relevant responses with \texttt{Gemini-3.1-Pro}~\citep{gemini31pro} and \texttt{GLM-5.2}~\citep{glm52}. Each judge scores every OE response in three independent calls that are then averaged (\S\ref{sec:method}). All open-weight models are self-deployed on an in-house H20 cluster; \texttt{Gemini-3.1-Pro} is the only closed-source judge model and is used only for the judge-robustness check of \mbox{\S\ref{sec:model-robust}}. Decoding settings, model sizes, and call counts are in \mbox{Appendix~\ref{sec:appendix-compute}}.
\section{Results}
\label{sec:results}
We first report how strongly each behavioral channel correlates with the self-report anchor \mbox{(\S\ref{sec:anchor-results})}, then compare self-anchored and reference-based results on PDB \mbox{(\S\ref{sec:ref-results})}, and finally check robustness across answering and judge models \mbox{(\S\ref{sec:model-robust})}.

\subsection{Correlation with the Self-Report Anchor}
\label{sec:anchor-results}
\label{sec:wclmsys-app}
\begin{figure}[tb]
  \centering
  \includegraphics[width=\linewidth]{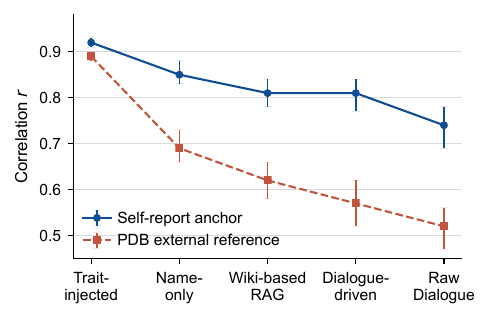}
  \caption{The two criteria order PDB's representation strategies the same way. Each point is a per-persona correlation (\mbox{\S\ref{sec:step3}}) Fisher-\mbox{$z$}-averaged over personas, channels and answering models, with by-persona cluster-bootstrap \mbox{$95\%$} CIs (\mbox{$3{,}000$} resamples). The self-anchored values are uniformly higher, but both curves decline monotonically as the representation supplies less explicit persona information.}
  \label{fig:strategy-two-criteria}
  \end{figure}

\paragraph{Overall level.}
Behavior tracks self-report closely on all three datasets. Per persona (Fisher-$z$-averaged over personas and the two channels), Pearson $r$ is \mbox{$0.74$--$0.92$} across PDB's five strategies (\mbox{Figure~\ref{fig:strategy-two-criteria}}) and across strategies and answering models on WildChat and LMSYS (\mbox{Table~\ref{tab:main-results}}). Per facet, the mean Pearson $r$ is \mbox{$0.80$}, \mbox{$0.79$} and \mbox{$0.69$} for SCN and \mbox{$0.71$}, \mbox{$0.67$} and \mbox{$0.59$} for OE on PDB, WildChat and LMSYS (\mbox{Table~\ref{tab:facets-anchor}}). The corresponding CCC means are lower, especially for OE (\mbox{$0.75$/$0.59$}, \mbox{$0.72$/$0.49$}, \mbox{$0.61$/$0.42$} for SCN/OE), reflecting mean shifts and scale compression (Appendix~\ref{sec:appendix-ccc}). SCN exceeds OE in every dataset $\times$ facet cell, as LLM-judge noise predicts.

\paragraph{Persona representation strategies.}
\mbox{Table~\ref{tab:main-results}} compares strategies across answering models. Profile-enriched representations (Dialogue-driven, USP Attributes-only, USP Profile) generally beat Raw Dialogue on the self-report criterion, but the best enriched strategy varies, so none is uniformly best. The largest Pearson gains are for \texttt{Hy3-preview} and \texttt{gemma-4-31B-it}, whose Raw-Dialogue $r$ is comparatively weak (\mbox{$0.74$} and \mbox{$0.79$} on WildChat, rising to \mbox{$0.88$} and \mbox{$0.89$}). With CIs (Figure~\ref{fig:strategy-by-model}), enrichment mainly makes the four models more alike: they span $r=0.74$--$0.88$ under Raw Dialogue on WildChat but fall within $0.06$ of one another under every enriched representation. On PDB the Pearson order is \emph{Trait-injected} (\mbox{$0.92$}) $>$ \emph{Name-only} (\mbox{$0.85$}) $>$ \emph{Wiki-based RAG} $\approx$ \emph{Dialogue-driven} (\mbox{$0.81$}) $>$ \emph{Raw Dialogue} (\mbox{$0.74$}). CCC is lower than Pearson and can reorder intermediate strategies---on LMSYS OE, persona-profile CCC vs.\ Pearson rankings yield Kendall $\tau=0$---but the headline gains remain positive: Trait-injected vs.\ Raw Dialogue on PDB, and USP Profile vs.\ Raw Dialogue on WildChat/LMSYS, in both channels (all six persona-bootstrap $95\%$ CIs exclude $0$; Appendix~\ref{sec:appendix-ccc}).

\begin{table}[t]
\centering\scriptsize
\setlength{\tabcolsep}{2.2pt}
\resizebox{\columnwidth}{!}{%
\begin{tabular}{l cccc}
\toprule
\textbf{Representation} & \texttt{GLM-5.1} & \texttt{Hy3} & \texttt{gemma-4} & \texttt{gpt-oss} \\
\midrule
\multicolumn{5}{l}{\emph{PDB, vs.\mbox{} external reference}} \\
Trait-injected      & \textbf{0.90}/0.80 & \textbf{0.91}/0.79 & \textbf{0.91}/0.80 & \textbf{0.84}/0.71 \\
Name-only           & 0.72/0.59 & 0.72/0.55 & 0.75/0.63 & 0.56/0.44 \\
Wiki-based RAG      & 0.65/0.51 & 0.62/0.47 & 0.68/0.54 & 0.53/0.40 \\
Dialogue-driven     & 0.53/0.41 & 0.61/0.46 & 0.58/0.44 & 0.56/0.41 \\
Raw Dialogue        & \underline{0.52}/0.40 & \underline{0.55}/0.38 & \underline{0.53}/0.40 & \underline{0.47}/0.36 \\
\addlinespace
\multicolumn{5}{l}{\emph{PDB, vs.\mbox{} self-report anchor}} \\
Trait-injected      & \textbf{0.93}/0.89 & \textbf{0.93}/0.88 & \textbf{0.93}/0.90 & \textbf{0.89}/0.82 \\
Name-only           & 0.86/0.78 & 0.85/0.74 & 0.88/0.79 & 0.83/0.71 \\
Wiki-based RAG      & 0.79/0.69 & 0.84/0.72 & 0.82/0.71 & 0.79/0.68 \\
Dialogue-driven     & 0.80/0.71 & 0.82/0.71 & 0.82/0.74 & 0.79/0.68 \\
Raw Dialogue        & \underline{0.75}/0.65 & \underline{0.75}/0.59 & \underline{0.78}/0.62 & \underline{0.65}/0.55 \\
\addlinespace
\multicolumn{5}{l}{\emph{WildChat, vs.\mbox{} self-report anchor}} \\
Dialogue-driven     & \textbf{0.90}/0.85 & \textbf{0.88}/0.80 & \textbf{0.89}/0.81 & 0.87/0.78 \\
USP Attributes-only & 0.88/0.81 & \textbf{0.88}/0.80 & 0.85/0.77 & 0.89/0.79 \\
USP Profile         & \underline{0.87}/0.81 & 0.85/0.75 & 0.86/0.80 & \textbf{0.90}/0.82 \\
Raw Dialogue        & 0.88/0.82 & \underline{0.74}/0.64 & \underline{0.79}/0.69 & \underline{0.84}/0.76 \\
\addlinespace
\multicolumn{5}{l}{\emph{LMSYS, vs.\mbox{} self-report anchor}} \\
Dialogue-driven     & \textbf{0.91}/0.86 & \textbf{0.88}/0.81 & 0.87/0.82 & 0.85/0.77 \\
USP Attributes-only & \textbf{0.91}/0.86 & 0.86/0.76 & \textbf{0.91}/0.84 & 0.91/0.81 \\
USP Profile         & \textbf{0.91}/0.86 & 0.87/0.78 & 0.88/0.82 & \textbf{0.92}/0.84 \\
Raw Dialogue        & \underline{0.88}/0.83 & \underline{0.79}/0.67 & \underline{0.77}/0.67 & \underline{0.84}/0.76 \\
\bottomrule
\end{tabular}%
}
\caption{Per-persona Pearson $r$ / Lin's CCC (\mbox{\S\ref{sec:step3}}) by strategy and answering model, Fisher-\mbox{$z$}-averaged over personas and the SCN/OE channels (default judge \texttt{DeepSeek-V4-Pro}). \textbf{Bold} and \underline{underlined} mark the strongest and weakest Pearson $r$ per model. CCC is the second number in each cell.}
\label{tab:main-results}
\end{table}

\paragraph{Facet-level results.}
\mbox{Table~\ref{tab:facets-anchor}} reports per-facet Pearson $r$ / CCC with the anchor. Respectfulness is among the strongest facets in every dataset $\times$ channel column (Pearson $r=0.67$--$0.88$). The weakest facet differs by dataset---Creative Imagination on PDB versus Productiveness on WildChat/LMSYS---and neither is the facet hardest to match against PDB's external reference (\mbox{\S\ref{sec:ref-results}}). OE is uniformly lower than SCN, but the two channels agree on within-dataset facet patterns. Both weak facets are strategy-dependent in ways this average conceals (Appendix~\ref{sec:appendix-facet-strategy}).

\begin{table}[t]
\centering\scriptsize
\setlength{\tabcolsep}{1.2pt}
\resizebox{\columnwidth}{!}{%
\begin{tabular}{l cc cc cc}
\toprule
& \multicolumn{2}{c}{PDB} & \multicolumn{2}{c}{WildChat} & \multicolumn{2}{c}{LMSYS} \\
\cmidrule(lr){2-3}\cmidrule(lr){4-5}\cmidrule(lr){6-7}
\textbf{Facet} & SCN & OE & SCN & OE & SCN & OE \\
\midrule
Assertiveness       & 0.84/0.78 & 0.75/0.69 & 0.75/0.43 & 0.64/0.57 & 0.63/0.39 & 0.52/0.45 \\
Respectfulness      & \textbf{0.86}/0.79 & \textbf{0.82}/0.71 & \textbf{0.88}/0.84 & \textbf{0.79}/0.43 & \textbf{0.78}/0.72 & \textbf{0.67}/0.35 \\
Productiveness      & 0.80/0.76 & 0.67/0.54 & \underline{0.69}/0.64 & \underline{0.54}/0.37 & \underline{0.53}/0.50 & \underline{0.51}/0.30 \\
Emot.\ Volatility   & 0.81/0.78 & 0.76/0.62 & 0.87/0.84 & 0.74/0.63 & 0.76/0.72 & 0.65/0.57 \\
Intell.\ Curiosity  & 0.79/0.75 & 0.65/0.56 & 0.70/0.62 & 0.59/0.42 & 0.68/0.58 & 0.56/0.39 \\
Creat.\ Imagination & \underline{0.67}/0.61 & \underline{0.52}/0.38 & 0.80/0.78 & 0.64/0.50 & 0.72/0.66 & 0.60/0.44 \\
\midrule
\textbf{Mean facet} & 0.80/0.75 & 0.71/0.59 & 0.79/0.72 & 0.67/0.49 & 0.69/0.61 & 0.59/0.42 \\
\bottomrule
\end{tabular}%
}
\caption{Per-facet Pearson $r$ / Lin's CCC with the self-report anchor (\mbox{\S\ref{sec:step3}}), computed across personas and Fisher-\mbox{$z$}-averaged over strategies and answering models. \textbf{Bold} and \underline{underlined} mark the strongest and weakest Pearson $r$ per column.}
\label{tab:facets-anchor}
\end{table}

\subsection{Relation to PDB's External Reference}
\label{sec:ref-results}
\label{sec:pdb-validation}
\paragraph{Correlation with the external reference is lower, and why.}
\mbox{Table~\ref{tab:facets-ref}} reports the per-facet correlation with PDB's external reference (same averaging as \mbox{Table~\ref{tab:facets-anchor}}; per-model version in Appendix~\ref{sec:appendix-recovery-full}). Averaged over facets it reaches \mbox{$0.62$} for self-report, \mbox{$0.57$} for SCN and \mbox{$0.52$} for OE, below the self-anchored correlations of \mbox{\S\ref{sec:anchor-results}}. Two mechanisms push in this direction and we cannot separate them. First, the reference is \emph{noisy and coarse}: community votes are Big Five \emph{domain} scores used here as domain-level proxies, not facet-level ground truth, and the two Open-Mindedness facets inherit one shared domain value (\mbox{Appendix~\ref{sec:appendix-pdb}}). Measurement error in either variable attenuates a correlation. Second, the anchor shares answering-model and prompt variance with the two behavioral channels, which inflates the self-anchored side. We therefore do not read the gap as evidence that the personas are inconsistent, nor the self-anchored numbers as an upper bound on fidelity to a real person. Respectfulness and Emotional Volatility are the most reliable facets against the reference, Assertiveness the hardest, and this ranking recurs across all four answering models (Appendix~\ref{sec:appendix-recovery-full}).

\begin{table}[t]
\centering\small
\setlength{\tabcolsep}{4pt}
\begin{tabular*}{\columnwidth}{@{\extracolsep{\fill}}l cc@{}}
\toprule
\textbf{Facet} & SCN & OE \\
\midrule
Assertiveness       & \underline{0.47} & \underline{0.43} \\
Respectfulness      & 0.66 & \textbf{0.67} \\
Productiveness      & 0.53 & 0.49 \\
Emot.\ Volatility   & \textbf{0.69} & 0.60 \\
Intell.\ Curiosity  & 0.50 & 0.46 \\
Creat.\ Imagination & 0.53 & \underline{0.42} \\
\midrule
\textbf{Mean facet} & 0.57 & 0.52 \\
\bottomrule
\end{tabular*}
\caption{Per-facet correlation of the two behavioral channels with PDB's external reference: Pearson \mbox{$r$} across the \mbox{$75$} personas, with the same averaging as \mbox{Table~\ref{tab:facets-anchor}}. The anchor's own accuracy against the reference is \mbox{Table~\ref{tab:anchor-vs-ref}}. \mbox{\textbf{Bold}} and \mbox{\underline{underlined}} mark the strongest and weakest facet per column.}
\label{tab:facets-ref}
\end{table}

\paragraph{How accurate the anchor itself is.}
The anchor correlates with the external reference slightly better than either behavioral channel \mbox{(\S\ref{sec:anchor-support}}, \mbox{Table~\ref{tab:anchor-vs-ref})} and ranks the facets the same way, so it is not the weak link in the chain. It is nonetheless far from an oracle, and its accuracy tracks how much persona information the representation supplies rather than any privilege of self-description: \mbox{$r=0.95$} under Trait-injected, which states the traits outright, against \mbox{$\approx\!0.30$} under the two dialogue-based strategies. That is the regime the conversation-derived datasets live in, and there all three profiles become indistinguishable (\mbox{$0.41$}/\mbox{$0.44$}/\mbox{$0.38$} for self-report/SCN/OE). We therefore cannot argue that the self-anchored correlation approximates the reference-based one merely because the anchor approximates the reference; that has to be established empirically, as the next paragraph does.

\paragraph{Self-anchored and reference-based criteria agree on strategy ordering.}
The framework is useful only if the self-anchored correlation, which needs no external reference, tracks the reference-based one. Aggregated per strategy (\mbox{Figure~\ref{fig:strategy-two-criteria}}), the two induce the same Pearson ordering: \emph{Trait-injected $>$ Name-only $>$ Wiki-based RAG $\approx$ Dialogue-driven $>$ Raw Dialogue}, with Trait-injected top-ranked in $71/75$ evaluated cases and Raw Dialogue consistently weakest. The self-anchored values are systematically higher, consistent with shared model and prompt variance.

Averaging over the 75 personas first can inflate this agreement, so we also ask, per persona $\times$ model $\times$ strategy cell: does closer agreement with the imperfect anchor coincide with closer agreement with the reference? The two remain positively associated (\mbox{$r=0.29$--$0.44$}), including cross-channel pairings and after excluding Trait-injected and Name-only (cross-channel $r=0.22$, $95\%$ CI $[0.14,0.30]$). This is \emph{criterion-related evidence} that the self-anchored correlation carries information about the reference-based one, not evidence that it is a proxy for human fidelity or downstream utility (\mbox{Appendix~\ref{sec:appendix-persona-robust}}). Exploratory splits of PDB's $40$ fictional and $35$ real personas show a larger real--fictional gap under Raw Dialogue than Trait-injected (SCN--anchor facet correlation; interaction $0.124$, unadjusted $95\%$ CI $[0.056,0.194]$; Appendix~\ref{sec:appendix-pdb}).

\subsection{Robustness Across Answering and Judge Models}
\label{sec:model-robust}
The representation-strategy patterns of \mbox{\S\ref{sec:anchor-results}--\S\ref{sec:ref-results}} hold across all four answering models, although intermediate-strategy order remains model-dependent (Table~\ref{tab:main-results}; CIs in Appendix~\ref{sec:appendix-model}). Re-scoring OE with two additional judges (Appendix~\ref{sec:appendix-judge-robust}) yields small aggregate differences (mean pairwise $|\Delta|=0.019$) and no qualitative ranking change.

\section{Discussion and Conclusion}
\label{sec:discussion}
We presented a self-anchored, cross-method \emph{internal-consistency diagnostic} for persona consistency when external user-level personality references are unavailable (\S\ref{sec:method}--\S\ref{sec:results}). A high correlation with the anchor indicates that a persona behaves as if it had a coherent trait profile across differently structured elicitation formats; on PDB it is also positively associated with the reference-based correlation \mbox{(\S\ref{sec:ref-results})}. It does not establish human fidelity or downstream utility. Profile-enriched representations show higher consistency than Raw Dialogue \mbox{(\S\ref{sec:anchor-results})}, with two non-exclusive readings: extraction may make scattered trait-relevant information easier to use, or it may discard individual-specific detail and yield a simpler persona that is easier to follow. Because we measure consistency rather than information preservation, higher consistency does not imply greater fidelity to the source conversation.

\typeout{DISCUSSION ENDS ON PAGE \thepage}

\section*{Limitations}
\typeout{LIMITATIONS STARTS ON PAGE \thepage}
\begin{itemize}
    \item \emph{Shared variance between the anchor and the behavioral channels.} All three profiles come from the same answering model, so shared response style may inflate agreement. MTMM (pass rates $68\%/75\%/63\%$ on PDB/WildChat/LMSYS; Appendix~\ref{sec:appendix-mtmm}) and persona-identifiability (Appendix~\ref{sec:appendix-identifiability}) support trait and persona specificity but cannot rule out shared-model bias. We interpret the metric comparatively, as an internal-consistency diagnostic rather than an absolute fidelity score. Eliciting the anchor from a separate model, or from the human user where available, would remove this source of shared variance.
    \item \emph{Non-independent repeated conditions.} Strategies, answering models and judges reuse the same personas. We treat the persona as the sampling unit for all CIs and permutation tests \mbox{(\S\ref{sec:step3})}.
    \item \emph{Noisy, domain-level external reference.} PDB annotations are community domain scores, not clinical or facet-level labels. Mapped values are domain-level proxies. Measurement error attenuates the reference-based correlations of \mbox{\S\ref{sec:ref-results}}. Removing either Open-Mindedness facet or pooling the two leaves Pearson strategy rankings unchanged (Appendix~\ref{sec:appendix-pdb}); that is a mapping-sensitivity check, not construct validation.
    \item \emph{Elicited, single-conversation behavior.} WildChat and LMSYS personas come from a single conversation, and all channels are elicited rather than observed. The diagnostic does not measure dynamic multi-turn assistant-task utility. Human--simulator validation remains future work.
    \item \emph{Facet, cultural and linguistic coverage.} The six-facet subset is selected for expected chat observability. Leave-one-facet-out checks support ranking robustness within this space (Appendix~\ref{sec:appendix-lofo}) but do not establish psychometric reliability or validity, nor coverage of personality in general. Datasets and prompts do not span cultural and linguistic variation in trait expression. Extending to all 15 BFI-2 facets and prospectively collected human self-reports remains future work.
\end{itemize}

\section*{Ethical Considerations}
This work infers personality-style profiles from user conversations, which raises several concerns. \emph{Privacy and consent.} The profiles are constructed from user--chatbot exchanges that users did not provide for the purpose of psychological assessment; we use WildChat and LMSYS only in their released, anonymized form. \emph{Profiling and manipulation.} We caution against deploying this framework to profile, target, or manipulate individuals, or to make consequential decisions about them; it is intended as a research tool for comparing persona representation strategies at the population level. \emph{Not a clinical instrument.} Inferred profiles can be inaccurate and should not be treated as reliable per-user assessments; the framework is \emph{not} a clinical or diagnostic instrument, and should not be used to make claims about any individual's actual personality (\S\ref{sec:discussion}). \emph{Aggregate reporting.} We report only aggregate, population-level statistics and do not release or discuss any individual-level inferred profile.

\emph{Artifacts and licensing.} The Software ZIP submitted with this paper contains the instruments, evaluation pipeline, aggregate results, README run instructions, 75 PDB personas $\times$ 5 strategies, and WildChat/LMSYS sample identifiers. We release under a permissive open-source license only the artifacts we created ourselves: the scenario and open-ended items, the judge template and rubric \mbox{(Appendix~\ref{sec:appendix-instruments})}, the elicitation and scoring pipeline code, and the aggregate results reported here, together with an example of raw model responses for one PDB persona across all five representation strategies, which involve no user data. Third-party material is not redistributed. We use WildChat under ODC-BY with attribution and LMSYS under its dataset license, which prohibits transferring the data to third parties, so for both we release only conversation identifiers, letting others rebuild our exact sample once the original terms are accepted. The PDB annotations are likewise not redistributed; we release the retrieval date (\S\ref{sec:setup}) and the code that maps the site's community votes to our external reference. The BFI-2 item text is not reproduced or released, as it is distributed by the instrument's authors \mbox{\citep{soto2017bfi2}}; we release only the facet assignment and reverse-keying of the items we use.

\bibliography{custom}

\clearpage

\appendix

\section{Elicitation Instruments and Judge Protocol}
\label{sec:appendix-judge}
\label{sec:appendix-instruments}
This appendix documents the three instruments a persona answers \mbox{(\S\ref{sec:step1}--\S\ref{sec:step2})} and the rules that turn their responses into facet scores, ending with the fixed template and rubric given to the LLM judge. \mbox{Table~\ref{tab:instruments}} summarizes the design. The complete item-by-item questionnaires and rubrics for all three channels are released with the code.

\begin{table}[h]
\centering\small
\setlength{\tabcolsep}{3pt}
\begin{tabular}{@{}l c c l@{}}
\toprule
\textbf{Channel} & \textbf{Items} & \textbf{Scored by} & \textbf{Repetition} \\
\midrule
Self-report & 18 & fixed key       & 5 shuffles, median \\
SCN         & 30 & option lookup   & single pass \\
OE          & 18 & LLM judge       & 3 calls, mean \\
\bottomrule
\end{tabular}
\caption{Design of the three elicitation instruments. Items are split evenly over the six retained facets ($3$, $5$ and $3$ items per facet, respectively), and every channel returns a score per facet on the same $1$--$5$ scale. Only the open-ended channel depends on a judge model.}
\label{tab:instruments}
\end{table}

\subsection{Self-Report Likert Questionnaire}
\label{sec:appendix-likert}
The self-anchored report uses 18 Likert items derived from the BFI-2 \mbox{\citep{soto2017bfi2}}, three per retained facet, presented on the instrument's five-point strongly-disagree to strongly-agree scale. Following standard BFI-2 scoring, items keyed against their facet are recoded so that a higher value always means more of the facet, and the three items of a facet are combined into one facet score on the $1$--$5$ scale. The questionnaire is administered five times per persona with the item order independently shuffled in each administration, and the reported facet score is the \emph{median} of the five administrations, which limits the influence of item-order effects and of a single anomalous run. Across the $9{,}000$ PDB facet scores ($75$ personas $\times$ $5$ strategies $\times$ $4$ answering models $\times$ $6$ facets), the five-administration range is zero for $61\%$ of scores and at most one-third of a scale point for $78\%$; the corresponding IQR is zero for $78\%$ and $\leq 1/3$ for $90\%$ (median range $=0$, mean range $=0.22$). Dispersion is therefore low for most evaluated PDB facet scores, though the maximum range reaches $2.67$, so the median does not hide variability in every cell. We do not reproduce the BFI-2 item text here, as it is distributed by the instrument's authors.

\subsection{Scenario-Based Questionnaire (SCN)}
\label{sec:appendix-scn}
The scenario channel uses 30 forced-choice situational items, five per facet. Each item describes an everyday situation and offers a set of mutually exclusive courses of action, and each option carries a score from $1$ to $5$ that was fixed when the item was written. A response is therefore scored by table lookup, with no judge model and no free-text interpretation, and the five item scores of a facet are combined into that facet's score. This is what makes the channel a useful counterpart to the open-ended one: its failure modes are those of item design rather than those of a judge.

\subsection{Open-Ended Prompts (OE)}
\label{sec:appendix-oe}
The open-ended channel uses 18 free-form prompts, three per facet, each asking the persona to recount concrete behavior rather than to rate itself. Because responses are unconstrained text, they cannot be scored by lookup: each response is scored from $1$ to $5$ by an LLM judge against an item-specific behavioral rubric, in three independent calls whose scores are averaged \mbox{(\S\ref{sec:appendix-judge-prompt}--\S\ref{sec:appendix-rubric})}. This is the only channel whose scores depend on a judge model, so its sensitivity to that choice is checked separately in \mbox{Appendix~\ref{sec:appendix-judge-robust}}.

\subsection{Judge Prompt Template}
\label{sec:appendix-judge-prompt}
Each open-ended answer is scored in three independent calls to the same judge model using the following fixed template (\texttt{\{question\}}, \texttt{\{answer\}}, \texttt{\{facet\_en\}}, \texttt{\{direction\}}, and \texttt{\{score\_anchors\_json\}} are substituted per item); the final item score is the arithmetic mean of the three returned scores.

\begin{promptbox}[title=Judge prompt template]
\footnotesize
You are a personality psychology expert evaluating an open-ended response. Based on the question, answer, and score anchors, assign a score from 1--5 (one decimal allowed). Briefly analyze first, then pick the best-matching anchor; cite 3--5 verbatim quotes from the answer as evidence.

\smallskip
\textbf{Question:} \texttt{\{question\}} \\
\textbf{Response:} \texttt{\{answer\}} \\
\textbf{Dimension:} \texttt{\{facet\_en\}} (\texttt{\{direction\}}) \\
\textbf{Scale:} 1--5 \\
\textbf{Score anchors:} \texttt{\{score\_anchors\_json\}}

\smallskip
\textbf{Scoring calibration} (applied in order): \emph{(1)} follow the scoring direction given in \texttt{\{direction\}} (e.g., for Emotional Volatility, higher score $=$ more volatile, i.e.\ less stable); \emph{(2)} vivid writing, wording, or length must not affect the score, only concrete behavior and specificity of experience; \emph{(3)} an answer matching only low anchors is scored 1--2 regardless of length or detail; \emph{(4)} actively look for evidence that the answer \emph{fails} to meet high-anchor requirements before defaulting to a mid-high score.

\smallskip
\textbf{Output JSON:} \texttt{\{"score": <float>, "evidence": [<str>, ...], "reasoning": <str>\}}
\end{promptbox}

\subsection{Scoring Rubric}
\label{sec:appendix-rubric}
Each of the 18 open-ended items (3 per facet) carries its own five-level behavioral rubric (one anchor per score $1$--$5$), used verbatim as \texttt{\{score\_anchors\_json\}} above. As an illustration, the rubric for one Assertiveness item is:

\begin{promptbox}[title=Rubric for one Assertiveness item]
\footnotesize
\emph{Direction:} higher $=$ more assertive.

\smallskip
\begin{tabular}{@{}r@{\ \ }p{0.82\columnwidth}@{}}
\textbf{1} & Almost never states their view; follows others throughout; willingly gives up control. \\[1pt]
\textbf{2} & Only hints indirectly; gives up if no one responds or after slight pushback. \\[1pt]
\textbf{3} & States their view but then leaves it to others; does not actively push or hold firm. \\[1pt]
\textbf{4} & States view clearly with reasons; pushes again when opposed, but may not lead all the way. \\[1pt]
\textbf{5} & Actively states and pushes others toward their preferred direction; holds and persuades under disagreement. \\
\end{tabular}
\end{promptbox}

The remaining 17 items follow the same schema (one behavioral anchor per score level, plus auxiliary high-/low-score signal lists used to disambiguate borderline cases); the full rubric set is released with the code.

\section{The 15 BFI-2 Facets}
\label{sec:appendix-bfi2}
Table~\ref{tab:bfi2-facets} lists the full BFI-2 facet inventory and marks the six facets retained by the criteria of \S\ref{sec:facets}.

\begin{table}[h]
\centering\small
\setlength{\tabcolsep}{5pt}
\begin{tabular}{l l c}
\toprule
\textbf{Domain} & \textbf{Facet} & \textbf{Retained} \\
\midrule
Extraversion          & Sociability            &            \\
                      & Assertiveness          & \checkmark \\
                      & Energy Level           &            \\
\addlinespace
Agreeableness         & Compassion             &            \\
                      & Respectfulness         & \checkmark \\
                      & Trust                  &            \\
\addlinespace
Conscientiousness     & Organization           &            \\
                      & Productiveness         & \checkmark \\
                      & Responsibility         &            \\
\addlinespace
Negative Emotionality & Anxiety                &            \\
                      & Depression             &            \\
                      & Emotional Volatility   & \checkmark \\
\addlinespace
Open-Mindedness       & Intellectual Curiosity & \checkmark \\
                      & Aesthetic Sensitivity  &            \\
                      & Creative Imagination   & \checkmark \\
\bottomrule
\end{tabular}
\caption{The 15 BFI-2 facets~\citep{soto2017bfi2}, grouped by domain. Check marks indicate the six facets retained in our chat-relevant subset (\S\ref{sec:facets}): one per domain, plus a second Open-Mindedness facet.}
\label{tab:bfi2-facets}
\end{table}

\section{PDB External-Reference Construction}
\label{sec:appendix-pdb}
PDB is the only dataset with an external personality reference, so it anchors the entire validation. This appendix documents how the reference is built, expanding the summary in \S\ref{sec:setup}.

\paragraph{Source annotations.} We use the community-aggregated Big Five (Five-Factor) domain scores from Personality Database (PDB), a public platform on which users vote on the personality of well-documented characters and public figures. For each persona these votes are aggregated into a score per Big Five domain.

\paragraph{Persona selection.} We take $75$ of the most-voted, best-documented personas ($40$ fictional characters and $35$ real public figures). Personas are chosen by documentation/vote volume rather than sampled at random, so the reference is drawn from the highest-agreement region of PDB. Main-text tables pool the two subsets; we report them separately below.

\paragraph{Confidence threshold.} To avoid low-agreement annotations, a Big Five domain is retained only if it is backed by at least $5$ community votes; personas without enough votes on the relevant domains are excluded before selection.

\paragraph{Mapping to the six BFI-2 facets.} PDB scores are at the Big Five \emph{domain} level, whereas our framework operates on six BFI-2 \emph{facets}. We use a fixed, manually specified mapping that follows the standard BFI-2 domain--facet hierarchy: Extraversion~$\rightarrow$~Assertiveness; Agreeableness~$\rightarrow$~Respectfulness; Conscientiousness~$\rightarrow$~Productiveness; Negative Emotionality~$\rightarrow$~Emotional Volatility; and Open-Mindedness~$\rightarrow$~Intellectual Curiosity \emph{and} Creative Imagination. Mapped scores are therefore domain-level \emph{proxies}, not facet-level ground truth. Because the reference is domain-level, the two Open-Mindedness facets inherit the same Open-Mindedness domain value; the external reference does not distinguish these two facets (the SCN/OE channels still elicit them separately). Removing either Open-Mindedness facet, or pooling the two into one averaged facet, leaves the Pearson strategy ranking against PDB unchanged in both behavioral channels (all Kendall $\tau=1$ on the $r_i$ ranking). Facet-difficulty order (Respectfulness / Emotional Volatility easiest, Assertiveness hardest) is likewise unchanged. This is a sensitivity check on duplicated mapping, not evidence of construct validity.

\paragraph{Scale.} The mapped scores are expressed on the same $1$--$5$ facet scale as the elicited profiles (stored per persona as \texttt{facet\_scores\_derived}) and used directly, with no further per-facet normalization. Because the reported statistic is a Pearson correlation \emph{across personas} (invariant to location and scale), the specific linear placement onto $1$--$5$ affects only the MAE/RMSE metrics (Table~\ref{tab:pdb-err}), not the correlations with the external reference.

\paragraph{Fictional vs.\ real-public-figure subsets.} Table~\ref{tab:pdb-subsets} reports SCN--anchor facet correlations ($r_f$) separately for the $40$ fictional characters and $35$ real public figures. Under Trait-injected the two subsets are indistinguishable ($0.92$ vs.\ $0.92$; difference CI includes $0$). Under Raw Dialogue the real--fictional gap is larger ($0.81$ vs.\ $0.68$). A difference-of-differences (Raw Dialogue gap minus Trait-injected gap) yields interaction $0.124$ (unadjusted persona-bootstrap $95\%$ CI $[0.056,0.194]$). The pattern is concentrated at the low-information end: when traits must be reconstructed from dialogue, real public figures are answered more consistently with their own self-report, plausibly because answering models have a stronger pretraining prior for widely documented people. The gap is a strategy $\times$ subset interaction, not a blanket subset main effect, and the comparison is exploratory (no multiplicity correction).

\begin{table}[h]
\centering\scriptsize
\setlength{\tabcolsep}{3pt}
\begin{tabular*}{\columnwidth}{@{\extracolsep{\fill}}l cc c}
\toprule
\textbf{Strategy} & Real ($n{=}35$) & Fict.\ ($n{=}40$) & Diff.\ [95\% CI] \\
\midrule
Trait-injected  & 0.92 & 0.92 & $0.00$ $[-0.03,0.03]$ \\
Name-only       & 0.79 & 0.77 & $0.02$ $[-0.04,0.08]$ \\
Wiki-based RAG  & 0.76 & 0.74 & $0.02$ $[-0.05,0.11]$ \\
Dialogue-driven & 0.80 & 0.74 & $0.06$ $[-0.02,0.14]$ \\
Raw Dialogue    & 0.81 & 0.68 & $0.13$ $[0.05,0.20]$ \\
\bottomrule
\end{tabular*}
\caption{PDB SCN--anchor per-facet Pearson $r$ ($r_f$), Fisher-$z$-averaged over facets and answering models, by fictional vs.\ real subset. Diff.\ is real minus fictional with a persona-clustered bootstrap $95\%$ CI.}
\label{tab:pdb-subsets}
\end{table}

\paragraph{Provenance of persona material.} The reference annotations are community Big Five votes as above. The persona-\emph{construction} inputs come from separate public sources that vary by strategy: Name-only supplies only the persona's name; Wiki-based RAG retrieves encyclopedic/biographical passages via the MediaWiki API; and Raw Dialogue and Dialogue-driven Free Description draw on verbatim quotation transcripts scraped from Wikiquote (the latter synthesizing a free-form description from the same transcripts). For Wiki-based RAG, Raw Dialogue, and Dialogue-driven Free Description, the persona's name (and, for RAG, the Wikipedia article title) is redacted from the retrieved/transcribed text via word-boundary string matching and replaced with a generic role placeholder before being shown to the answering model, so these three strategies never reveal the persona's actual name. Trait-injected instead injects an explicit trait description directly, as an oracle-style condition (\S\ref{sec:setup}); like Name-only, it does reveal the persona's name. Except for Trait-injected, none of these construction sources overlap with the community Big Five reference annotations, so recovery measures agreement between the community reference and a profile elicited from the constructed persona.

\section{Persona Representation Strategies (Detail)}
\label{sec:appendix-construction}
This appendix expands the strategy summary in \S\ref{sec:construction}.

\paragraph{Shared conversation-based strategies.} Raw Dialogue supplies the source conversation directly as context, following the in-context-history strategy common in role-playing language agents~\citep{chen2024persona}; Dialogue-driven Free Description instead supplies a free-form description synthesized from that same conversation by \texttt{GPT-5.5}, in the spirit of dialogue-history-based persona characterization without a predefined profile~\citep{cheng-etal-2024-dialogues,wang-etal-2025-know}. On PDB, the source consists of reference dialogue transcripts (\S\ref{sec:appendix-pdb}); on WildChat/LMSYS, it consists of one complete user--chatbot conversation (\S\ref{sec:setup}). On PDB, both strategies redact the persona's name from the underlying transcript, replacing it with a generic role placeholder (\emph{this character}/\emph{this public figure}).

\paragraph{Dataset-specific strategies.} PDB additionally supports three strategies not available for anonymous conversation-derived personas: \emph{Name-only} (only the persona's name is given, with no other material), \emph{Wiki-based RAG} (retrieved encyclopedic passages about the persona, with the persona's name and article title likewise redacted and replaced by the generic role placeholder, following retrieval-grounded role-profile construction~\citep{wang-etal-2024-rolellm}), and \emph{Trait-injected} (the persona's name plus an explicit trait profile injected into the prompt, an oracle-style condition with direct trait information). Name-only and Trait-injected are thus the only PDB strategies that reveal the persona's actual name. WildChat and LMSYS additionally use two strategies produced by the USP profile-extraction pipeline~\citep{wang-etal-2025-know}: \emph{USP Attributes-only} (the extracted attribute representation without its synthesized free-form profile) and \emph{USP Profile} (the complete synthesized profile); we run the pipeline with \texttt{GPT-5.5} as the extractor. All three synthesized representations are therefore generated once, before evaluation, and held fixed across the four answering models, so differences between answering models are not confounded with differences in the representation text.

\paragraph{How much text each strategy supplies.} \mbox{Table~\ref{tab:input-length}} reports the length of the persona representation that the answering model receives, which varies by more than two orders of magnitude across strategies. Length alone does not determine consistency: Wiki-based RAG supplies by far the longest input yet ranks below the two shortest strategies, Name-only and Trait-injected \mbox{(Figure~\ref{fig:strategy-two-criteria})}, so what matters is how explicitly the input states trait-relevant information rather than how much text it contains.

\begin{table}[t]
\centering\small
\setlength{\tabcolsep}{3pt}
\begin{tabular*}{\columnwidth}{@{\extracolsep{\fill}}l rr@{}}
\toprule
\textbf{Strategy} & \textbf{Mean} & \textbf{Median} \\
\midrule
\multicolumn{3}{@{}l}{\emph{PDB}} \\
\quad Name-only            & $29$     & $28$ \\
\quad Trait-injected       & $364$    & $364$ \\
\quad Dialogue-driven      & $590$    & $590$ \\
\quad Raw Dialogue         & $2{,}325$  & $2{,}652$ \\
\quad Wiki-based RAG       & $11{,}683$ & $7{,}524$ \\
\addlinespace
\multicolumn{3}{@{}l}{\emph{WildChat}} \\
\quad USP Profile          & $647$    & $630$ \\
\quad USP Attributes-only  & $1{,}195$  & $1{,}101$ \\
\addlinespace
\multicolumn{3}{@{}l}{\emph{LMSYS}} \\
\quad USP Profile          & $753$    & $720$ \\
\quad USP Attributes-only  & $1{,}431$  & $1{,}413$ \\
\bottomrule
\end{tabular*}
\caption{Approximate length, in tokens, of the persona representation supplied to the answering model, excluding the instrument text. Counts are estimated from character length divided by four rather than measured with a tokenizer.}
\label{tab:input-length}
\end{table}

\section{Additional Result Tables}
\label{sec:appendix-tables}
All averaged correlations in this appendix use Fisher-$z$ aggregation, matching the main text; open-ended scores use the default judge \texttt{DeepSeek-V4-Pro} unless noted. Higher $r$ is better throughout. For brevity we write \emph{ext.-ref.\mbox{} correlation} for the correlation of a profile with the external reference and \emph{self-report correlation} for its correlation with the self-report anchor \mbox{(\S\ref{sec:step3})}.

\subsection{Confidence Intervals for the Per-Facet Correlations}
\label{sec:appendix-facet-cis}
\mbox{Tables~\ref{tab:facets-anchor}} and~\mbox{\ref{tab:facets-ref}} report point estimates only. \mbox{Table~\ref{tab:facet-cis}} adds the intervals: analytic Fisher-\mbox{$z$} \mbox{$95\%$} CIs for the external-reference correlations, which are single across-persona correlations at \mbox{$n=75$}, and the CI of the column mean for the self-report correlations, which average over the \mbox{$20$} (PDB) or \mbox{$16$} (WildChat, LMSYS) strategy \mbox{$\times$} answering-model cells. The intervals are wide relative to the differences between facets, so we read the facet \emph{ordering} rather than individual gaps.

\begin{table}[h]
\centering\scriptsize
\setlength{\tabcolsep}{2.5pt}
\begin{tabular}{l ccc}
\toprule
 & \multicolumn{3}{c}{\textbf{Correlation with ext.\mbox{} reference}} \\
\cmidrule(lr){2-4}
\textbf{Facet} & Self-report & SCN & OE \\
\midrule
Assertiveness          & 0.49 [.30,.65] & 0.47 [.27,.63] & 0.43 [.23,.60] \\
Respectfulness         & 0.71 [.57,.80] & 0.66 [.50,.77] & 0.67 [.52,.78] \\
Productiveness         & 0.66 [.51,.77] & 0.53 [.35,.68] & 0.49 [.30,.65] \\
Emot.\ Volatility      & 0.70 [.56,.80] & 0.69 [.54,.79] & 0.60 [.43,.73] \\
Intell.\ Curiosity     & 0.58 [.41,.72] & 0.50 [.31,.65] & 0.46 [.26,.62] \\
Creat.\ Imagination    & 0.55 [.37,.69] & 0.53 [.35,.68] & 0.42 [.21,.59] \\
\addlinespace
\toprule
 & \multicolumn{3}{c}{\textbf{Correlation with self-report anchor}} \\
\cmidrule(lr){2-4}
\textbf{Channel} & PDB & WildChat & LMSYS \\
\midrule
SCN & 0.80 [.70,.87] & 0.79 [.71,.86] & 0.69 [.58,.78] \\
OE  & 0.71 [.57,.81] & 0.67 [.54,.76] & 0.59 [.44,.70] \\
\bottomrule
\end{tabular}
\caption{Intervals for \mbox{Tables~\ref{tab:facets-anchor}} and~\mbox{\ref{tab:facets-ref}}. Top: per-facet analytic Fisher-\mbox{$z$} \mbox{$95\%$} CIs for the correlations with PDB's external reference. Bottom: the facet-mean self-report correlation per dataset and channel, with the \mbox{$95\%$} CI of that mean.}
\label{tab:facet-cis}
\end{table}

\subsection{Full Per-Answering-Model Ext.-Ref.\ Correlation (PDB)}
\label{sec:appendix-recovery-full}
Table~\ref{tab:recovery-full} expands \mbox{Table~\ref{tab:facets-ref}} to all four answering models. The facet ordering (Respectfulness/Emotional Volatility highest, Assertiveness lowest) recurs across every model.

\begin{table*}[t]
\centering\small
\setlength{\tabcolsep}{3.5pt}
\begin{tabular}{l cc cc cc cc}
\toprule
 & \multicolumn{2}{c}{\textbf{\texttt{GLM-5.1-FP8}}}
 & \multicolumn{2}{c}{\textbf{\texttt{Hy3-preview}}}
 & \multicolumn{2}{c}{\textbf{\texttt{gemma-4-31B-it}}}
 & \multicolumn{2}{c}{\textbf{\texttt{gpt-oss-120b}}} \\
\cmidrule(lr){2-3}\cmidrule(lr){4-5}\cmidrule(lr){6-7}\cmidrule(lr){8-9}
\textbf{Facet} & SCN & OE & SCN & OE & SCN & OE & SCN & OE \\
\midrule
Assertiveness          & 0.47 & 0.41 & 0.49 & 0.41 & 0.52 & 0.43 & 0.40 & 0.47 \\
Respectfulness         & 0.65 & 0.69 & 0.69 & 0.70 & 0.69 & 0.67 & 0.58 & 0.60 \\
Productiveness         & 0.59 & 0.52 & 0.52 & 0.54 & 0.56 & 0.53 & 0.45 & 0.36 \\
Emotional Volatility   & 0.71 & 0.64 & 0.70 & 0.60 & 0.67 & 0.63 & 0.66 & 0.52 \\
Intellectual Curiosity & 0.57 & 0.47 & 0.48 & 0.42 & 0.48 & 0.45 & 0.45 & 0.49 \\
Creative Imagination   & 0.52 & 0.52 & 0.53 & 0.43 & 0.60 & 0.43 & 0.48 & 0.26 \\
\midrule
\textbf{Mean facet}    & 0.59 & 0.55 & 0.57 & 0.53 & 0.59 & 0.53 & 0.51 & 0.46 \\
\bottomrule
\end{tabular}
\caption{Full per-answering-model correlation with the external reference on PDB: per-facet Pearson $r$ across personas between the PDB reference and the SCN or OE profiles, averaged over the five representation strategies in Fisher-$z$ space (pooled over fictional-character and real-public-figure subsets). This expands the two behavior-oriented columns of \mbox{Table~\ref{tab:facets-ref}}; the self-report column of that table is not broken out by model.}
\label{tab:recovery-full}
\end{table*}

\subsection{Per-Facet Self-Report Correlation by Representation Strategy}
\label{sec:appendix-facet-strategy}
\begin{figure*}[t]
\centering
\includegraphics[width=\textwidth]{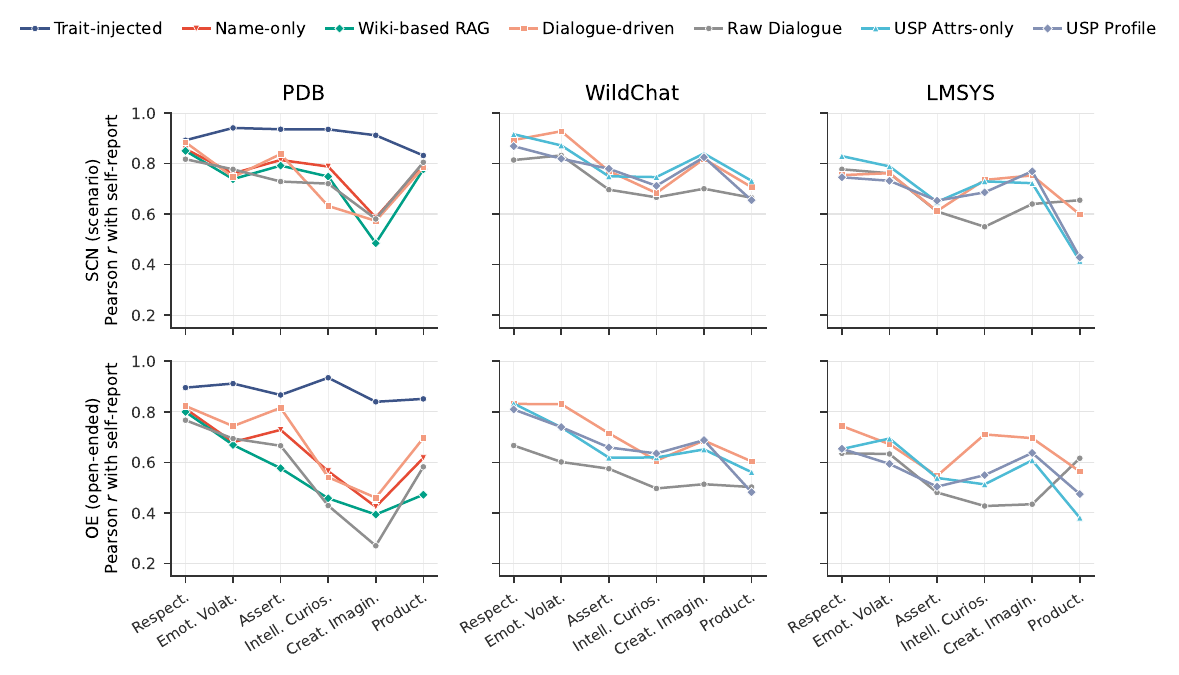}
\caption{Correlation with the self-report anchor, per facet and persona representation strategy, by dataset (columns) and elicitation channel (rows); the strategy-averaged values are those of \mbox{Table~\ref{tab:facets-anchor}}. Values are Pearson $r$ across personas, Fisher-$z$-averaged over the four answering models, with the default judge \texttt{DeepSeek-V4-Pro} for OE. Facets are ordered by their mean correlation across all six dataset $\times$ channel conditions and all six panels share one $r$ axis. A wide fan of lines marks a facet whose main-text value averages over qualitatively different conditions: PDB's Creative Imagination (oracle Trait-injected far above the rest) and LMSYS's Productiveness (the two USP-derived strategies far below the two dialogue-based ones) are the two clearest cases.}
\label{fig:convergence-strategy}
\end{figure*}

\mbox{Table~\ref{tab:facets-anchor}} averages over the $20$ (PDB) or $16$ (WildChat, LMSYS) strategy $\times$ answering-model cells. That is the right summary for a per-facet claim, but it can hide facet $\times$ strategy interactions. Figure~\ref{fig:convergence-strategy} undoes the strategy average.

Most of the averaged picture survives: Respectfulness is among the strongest facets under nearly every strategy, and every strategy shows the same OE-below-SCN offset. Two interactions do not survive, and each qualifies a claim that the averaged view invites.

First, PDB's weak Creative Imagination is almost entirely an oracle-versus-rest split. It correlates with the anchor at $r=0.91$ (SCN) under Trait-injected, which supplies the trait profile directly, but at only $0.48$--$0.59$ under the four strategies that do not. This is the widest strategy spread of any PDB facet ($0.43$ in SCN, $0.57$ in OE). The averaged $0.67$ therefore \emph{understates} the problem: outside the oracle condition, Creative Imagination is the one facet on which a PDB persona does not behave consistently with its own self-report, which is what the channel-observability argument of \S\ref{sec:facets} predicts for a facet whose triggering topics are scarce in dialogue.

Second, LMSYS's weak Productiveness is specific to the USP-derived strategies: $r=0.66$ and $0.60$ under Raw Dialogue and Dialogue-driven Free Description, against $0.41$ and $0.43$ under USP Attributes-only and USP Profile. Read per strategy, Productiveness is not a uniformly weak facet on LMSYS but one that the USP extraction step degrades, so the dataset-level statement in \mbox{\S\ref{sec:anchor-results}} is a property of the strategy mix rather than of the dataset.

The interaction is itself dataset-dependent: WildChat is comparatively flat (per-facet strategy spreads of $0.08$--$0.14$ in SCN), whereas PDB and LMSYS contain the two cases above. We therefore report the averaged view in the main text, but do not read individual facet values as strategy-independent properties of a dataset.

\subsection{Answering-Model Breakdown with Confidence Intervals}
\label{sec:appendix-model}
\begin{figure*}[t]
\centering
\includegraphics[width=\textwidth]{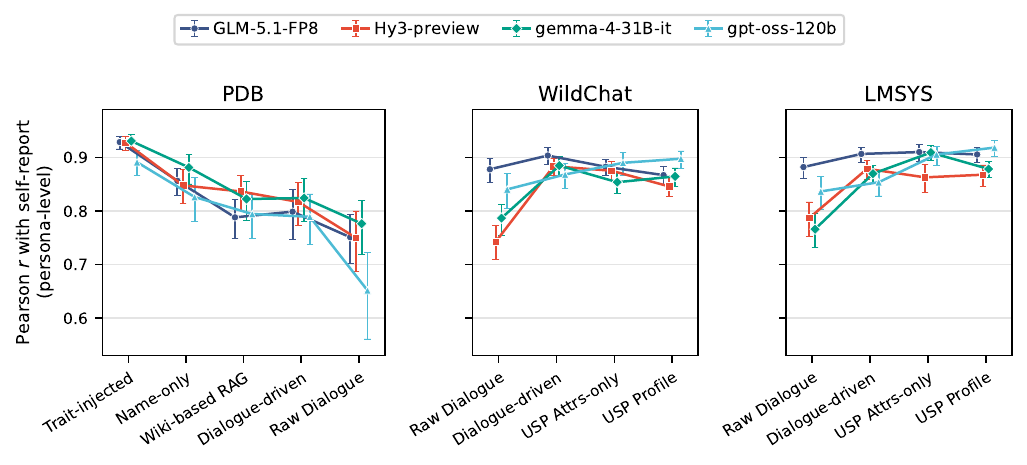}
\caption{Correlation with the self-report anchor by persona representation strategy and answering model, per dataset; the numeric form is \mbox{Table~\ref{tab:main-results}}. Values are per-persona correlations (\mbox{\S\ref{sec:step3}}), Fisher-$z$-averaged over personas and the SCN/OE channels with the default judge for OE. Error bars are by-persona cluster-bootstrap 95\% CIs ($4{,}000$ resamples; $n=75$ personas for PDB, $100$ for WildChat and LMSYS) and points are dodged horizontally to keep them legible. All three panels share one $r$ axis. Adjacent enriched strategies are not separated by their intervals, whereas the model spread under Raw Dialogue is.}
\label{fig:strategy-by-model}
\end{figure*}

The per-answering-model numbers for all three datasets are given in the main text \mbox{(Table~\ref{tab:main-results})}. \mbox{Figure~\ref{fig:strategy-by-model}} plots their self-report half against a common axis and adds the by-persona bootstrap CIs that the table omits at this level of disaggregation.

Read with the intervals, the three datasets tell one story. Under every enriched representation the four answering models are separated by at most $0.06$, and their CIs largely overlap, so the identity of the best strategy in a given column of \mbox{Table~\ref{tab:main-results}} should not be over-interpreted. Under Raw Dialogue the models separate by $0.12$--$0.14$ on all three datasets, and there the intervals genuinely come apart: on WildChat, \texttt{GLM-5.1-FP8} at $0.88$ $[0.85,0.90]$ against \texttt{Hy3-preview} at $0.74$ $[0.71,0.77]$. What an enriched representation buys is therefore less a uniform gain than a reduction in how much the answering model matters. PDB is the exception that fits: because its strategies vary how much external information is injected rather than how conversation is summarized, its profile keeps declining across the whole strategy axis instead of flattening after the first enrichment step.

\subsection{Judge-Model Robustness (Full Comparison)}
\label{sec:appendix-judge-robust}
Main-text results use \texttt{DeepSeek-V4-Pro} as the judge throughout. As a robustness check, Table~\ref{tab:judges} additionally re-scores the relevant open-ended responses with \texttt{Gemini-3.1-Pro} and \texttt{GLM-5.2}. The additional judges re-score all PDB open-ended responses and the two dialogue-based strategies on WildChat and LMSYS; the USP-derived strategies are excluded from this comparison because \texttt{Gemini-3.1-Pro} was not run on them (see \S\ref{sec:setup}). The aggregate correlations remain broadly consistent: the mean pairwise absolute difference across the reported dataset$\times$metric cells is $0.019$ (largest pairwise mean: \texttt{DeepSeek-V4-Pro} vs.\ \texttt{GLM-5.2}, $0.024$), and no qualitative ranking conclusion changes.

\begin{table}[t]
\centering\small
\setlength{\tabcolsep}{2pt}
\begin{tabular}{l cccc}
\toprule
& \multicolumn{2}{c}{\textbf{PDB}} & & \\
\cmidrule(lr){2-3}
\textbf{Judge} & Ext.\mbox{} ref. & Self-rep. & WildChat & LMSYS \\
\midrule
\texttt{DeepSeek-V4-Pro} & 0.598 & 0.712 & 0.736 & 0.750 \\
\texttt{Gemini-3.1-Pro}  & 0.581 & 0.709 & 0.766 & 0.796 \\
\texttt{GLM-5.2}         & 0.583 & 0.690 & 0.760 & 0.786 \\
\bottomrule
\end{tabular}
\caption{Open-ended results under three judge models. PDB columns report the correlation with the external reference and with the self-report anchor; WildChat and LMSYS report the self-report correlation for the two dialogue-based strategies only, because the USP-derived strategies were not run with all three judges. Each entry is a single pooled correlation from the judge-comparison pipeline, so the absolute values are not directly comparable to the per-facet Fisher-$z$ means of \mbox{Tables~\ref{tab:facets-anchor}} and~\mbox{\ref{tab:facets-ref}}; only the (aggregation-invariant) cross-judge differences are interpreted. Aggregate differences are small (mean pairwise $|\Delta|=0.019$), and the qualitative conclusions are stable across judges.}
\label{tab:judges}
\end{table}

\subsection{Per-Facet Error Metrics (PDB, vs.\ External Reference)}
\label{sec:appendix-error-metrics}
Table~\ref{tab:pdb-err} reports MAE and RMSE (as \texttt{MAE\,/\,RMSE}) on the raw facet-score scale ($1$--$5$), computed with the same pooling as \mbox{Table~\ref{tab:facets-ref}} (pooled over the fictional-character and real-public-figure subsets, averaged over the five PDB representation strategies; open-ended scores use the default judge, \texttt{DeepSeek-V4-Pro}). Lower is better. The facet ordering by error magnitude is consistent with the Pearson-based ordering there: Respectfulness and Emotional Volatility have the lowest error, Assertiveness and Productiveness the highest.

\begin{table*}[t]
\centering\small
\setlength{\tabcolsep}{3pt}
\begin{tabular}{l cc cc cc cc}
\toprule
 & \multicolumn{2}{c}{\textbf{\texttt{GLM-5.1-FP8}}}
 & \multicolumn{2}{c}{\textbf{\texttt{Hy3-preview}}}
 & \multicolumn{2}{c}{\textbf{\texttt{gemma-4-31B-it}}}
 & \multicolumn{2}{c}{\textbf{\texttt{gpt-oss-120b}}} \\
\cmidrule(lr){2-3}\cmidrule(lr){4-5}\cmidrule(lr){6-7}\cmidrule(lr){8-9}
\textbf{Facet} & SCN & OE & SCN & OE & SCN & OE & SCN & OE \\
\midrule
Assertiveness          & 1.08\,/\,1.31 & 0.99\,/\,1.23 & 1.20\,/\,1.45 & 1.00\,/\,1.22 & 1.11\,/\,1.36 & 1.05\,/\,1.29 & 1.13\,/\,1.37 & 0.81\,/\,1.00 \\
Respectfulness         & 1.09\,/\,1.27 & 0.64\,/\,0.83 & 0.87\,/\,1.06 & 0.66\,/\,0.86 & 0.76\,/\,1.00 & 0.76\,/\,0.98 & 1.09\,/\,1.32 & 0.69\,/\,0.90 \\
Productiveness         & 0.92\,/\,1.15 & 0.73\,/\,0.93 & 0.98\,/\,1.20 & 0.77\,/\,0.97 & 1.01\,/\,1.24 & 0.82\,/\,1.00 & 1.18\,/\,1.43 & 1.15\,/\,1.41 \\
Emotional Volatility   & 0.77\,/\,0.95 & 0.76\,/\,0.95 & 0.73\,/\,0.92 & 0.87\,/\,1.09 & 0.79\,/\,0.99 & 0.74\,/\,0.95 & 0.85\,/\,1.06 & 0.96\,/\,1.18 \\
Intellectual Curiosity & 0.75\,/\,0.94 & 0.84\,/\,1.04 & 0.84\,/\,1.06 & 0.83\,/\,1.04 & 0.83\,/\,1.06 & 0.83\,/\,1.05 & 0.79\,/\,1.01 & 0.86\,/\,1.12 \\
Creative Imagination   & 0.77\,/\,0.97 & 0.71\,/\,0.90 & 0.87\,/\,1.11 & 0.80\,/\,0.96 & 0.73\,/\,0.93 & 0.75\,/\,0.93 & 0.81\,/\,1.05 & 0.78\,/\,0.97 \\
\midrule
\textbf{Mean facet}    & 0.89\,/\,1.10 & 0.78\,/\,0.98 & 0.91\,/\,1.13 & 0.82\,/\,1.02 & 0.87\,/\,1.10 & 0.83\,/\,1.03 & 0.98\,/\,1.21 & 0.88\,/\,1.10 \\
\bottomrule
\end{tabular}
\caption{Deviation from the external reference on PDB: per-facet error as \texttt{MAE\,/\,RMSE} (scale $1$--$5$; lower is better) between the PDB reference annotations and the scenario-based (SCN) or open-ended (OE) profiles. Same pooling and default judge as \mbox{Table~\ref{tab:facets-ref}}. }
\label{tab:pdb-err}
\end{table*}

\subsection{Persona-Level Robustness of the Two Correlations (PDB)}
\label{sec:appendix-persona-robust}
This appendix reports the persona-level analysis summarized in \mbox{\S\ref{sec:ref-results}} (criterion-related validity). Averaging over the $75$ personas, as \mbox{Figure~\ref{fig:strategy-two-criteria}} does, cancels persona-level noise and can inflate the agreement between the two correlations, so here we keep every persona $\times$ answering-model $\times$ representation-strategy cell separate, with SCN and OE analyzed individually ($n=1500$ per channel). \mbox{Table~\ref{tab:persona-level}} gives the compact summary: the two correlations remain positively associated for both channels, including after excluding Trait-injected and Name-only, the two strategies that most directly inject external information, and including when the two are drawn from \emph{different} elicitation channels (rows \emph{SCN$\to$OE} and \emph{OE$\to$SCN}), which rules out same-channel measurement overlap as the sole explanation: cross-channel associations stay positive (\mbox{$r=0.31$--$0.32$} overall and \mbox{$r=0.22$} after the exclusions), although weaker than the matched same-channel pairings (\mbox{$r=0.39$--$0.44$} overall). Table~\ref{tab:persona-robust} reports the full analysis over the same cells.

\begin{table}[t]
\centering\small
\setlength{\tabcolsep}{4pt}
\begin{tabular}{l cc}
\toprule
\textbf{Pairing} & \textbf{All strategies} & \textbf{Excl.\ Trait+Name} \\
\midrule
SCN$\to$SCN & 0.39 [0.34, 0.45] & 0.29 [0.22, 0.37] \\
OE$\to$OE   & 0.44 [0.38, 0.50] & 0.34 [0.26, 0.42] \\
SCN$\to$OE  & 0.32 [0.25, 0.38] & 0.22 [0.14, 0.30] \\
OE$\to$SCN  & 0.31 [0.25, 0.36] & 0.22 [0.14, 0.30] \\
\bottomrule
\end{tabular}
\caption{Persona-level association between the correlation with the external reference and the correlation with the self-report anchor on PDB, by channel pairing: the first channel supplies the external-reference correlation and the second the anchor correlation (e.g.\mbox{} \emph{SCN$\to$OE} pairs the SCN--reference correlation with the OE--anchor correlation). Brackets denote persona-cluster bootstrap 95\% CIs; full robustness battery in Table~\ref{tab:persona-robust}.}
\label{tab:persona-level}
\end{table}

Table~\ref{tab:persona-robust} also asks whether the link between the two correlations is specific to comparing a channel with itself. The \emph{Pooled}, \emph{SCN}, and \emph{OE} rows always pair a channel's ext.-ref.\mbox{} correlation with the \emph{same} channel's self-report correlation, so part of the association could in principle reflect variance shared by using one elicitation channel on both axes (e.g.\ shared judge or prompt-format artifacts) rather than a genuine cross-method signal. We therefore add two \emph{cross}-channel pairings that never share a channel across the two axes: the SCN ext.-ref.\mbox{} correlation against the OE self-report correlation, and the OE ext.-ref.\mbox{} correlation against the SCN self-report correlation (rows \emph{SCN$\to$OE} and \emph{OE$\to$SCN}). Both quantities in every pairing are six-facet Pearson correlations already present in the underlying persona-level data, so this check required no new model or judge calls, only recombining existing columns.

For every one of the $3$ strategy subsets $\times$ $5$ channel pairings ($15$ cells: same-channel \emph{Pooled}/\emph{SCN}/\emph{OE}, plus cross-channel \emph{SCN$\to$OE} and \emph{OE$\to$SCN}), Table~\ref{tab:persona-robust} reports: the pooled persona-level Pearson $r$; a $95\%$ confidence interval from a cluster bootstrap that resamples \emph{personas} (not rows) with replacement, since each persona contributes up to $20$ non-independent rows per pairing; the fixed-effect coefficient on the ext.-ref.\mbox{} correlation from a mixed-effects regression of the self-report correlation on it, with a random intercept per persona and fixed effects for strategy and answering model (only the \emph{Pooled} row additionally controls for channel, since it is the only row that mixes both SCN and OE on the same axis; the \emph{SCN}, \emph{OE}, \emph{SCN$\to$OE}, and \emph{OE$\to$SCN} rows each fix a single channel pairing, leaving no channel variation to control for; all $15$ coefficients are significant at $p<10^{-3}$, with the weakest at $p=8.7\times10^{-4}$ for \emph{OE$\to$SCN}/Excl.\ Trait+Name); and the range of $r$ across four leave-one-answering-model-out refits. Every cut yields a positive, statistically significant relationship. The cross-channel pairings are weaker than the same-channel ones ($r{\approx}0.22$--$0.32$ vs.\ $0.29$--$0.44$; mixed $\beta$ roughly halves), indicating that part of the same-channel association reflects variance shared by using one elicitation channel on both axes. The cross-channel relationship nonetheless remains positive and significant throughout, so the link between the two correlations is not solely an artifact of comparing a channel with itself. The leave-one-model-out range never exceeds $0.05$ in any row, indicating none of the fifteen cells is driven by a single answering model.

\begin{table}[t]
\centering\footnotesize
\setlength{\tabcolsep}{3pt}
\begin{tabular*}{\columnwidth}{@{\extracolsep{\fill}}l c c c@{}}
\toprule
\textbf{Channel pairing} & $r$ [95\% CI] & $\beta$ & LOMO \\
\midrule
\multicolumn{4}{@{}l}{\emph{All strategies} ($n=1500$; $3000$ pooled)} \\
\quad Pooled (same-ch.)  & 0.42 [0.36, 0.46] & 0.27 & [0.40, 0.42] \\
\quad SCN                & 0.39 [0.34, 0.45] & 0.22 & [0.37, 0.42] \\
\quad OE                 & 0.44 [0.38, 0.50] & 0.32 & [0.43, 0.45] \\
\quad SCN$\to$OE (cross) & 0.32 [0.25, 0.38] & 0.17 & [0.29, 0.33] \\
\quad OE$\to$SCN (cross) & 0.31 [0.25, 0.36] & 0.14 & [0.30, 0.32] \\
\addlinespace
\multicolumn{4}{@{}l}{\emph{Excl.\ Trait} ($n=1200$; $2400$ pooled)} \\
\quad Pooled (same-ch.)  & 0.33 [0.28, 0.39] & 0.24 & [0.32, 0.34] \\
\quad SCN                & 0.31 [0.24, 0.38] & 0.20 & [0.29, 0.34] \\
\quad OE                 & 0.36 [0.29, 0.44] & 0.28 & [0.35, 0.37] \\
\quad SCN$\to$OE (cross) & 0.23 [0.16, 0.31] & 0.17 & [0.20, 0.25] \\
\quad OE$\to$SCN (cross) & 0.22 [0.15, 0.28] & 0.11 & [0.21, 0.24] \\
\addlinespace
\multicolumn{4}{@{}l}{\emph{Excl.\ Trait$+$Name} ($n=900$; $1800$ pooled)} \\
\quad Pooled (same-ch.)  & 0.31 [0.25, 0.37] & 0.22 & [0.29, 0.32] \\
\quad SCN                & 0.29 [0.22, 0.37] & 0.18 & [0.26, 0.32] \\
\quad OE                 & 0.34 [0.26, 0.42] & 0.25 & [0.32, 0.34] \\
\quad SCN$\to$OE (cross) & 0.22 [0.14, 0.30] & 0.15 & [0.18, 0.25] \\
\quad OE$\to$SCN (cross) & 0.22 [0.14, 0.30] & 0.10 & [0.20, 0.24] \\
\bottomrule
\end{tabular*}
\caption{Persona-level robustness of the association between the ext.-ref.\mbox{} correlation and the self-report correlation on PDB (no averaging over personas). ``Excl.\ Trait'' drops the Trait-injected strategy; ``Excl.\ Trait$+$Name'' additionally drops Name-only. $95\%$ CI is a by-persona cluster bootstrap ($5{,}000$ resamples). $\beta$ is the ext.-ref.-correlation fixed-effect coefficient from a mixed-effects regression with a random intercept per persona and fixed effects for strategy and answering model (the \emph{Pooled} row additionally adds a channel fixed effect, since it is the only row mixing SCN and OE; not shown). All $15$ coefficients are significant at $p<10^{-3}$ (not shown). LOMO is the range of $r$ over four leave-one-answering-model-out refits, which varies by at most $0.05$ within every row.}
\label{tab:persona-robust}
\end{table}

\subsection{Leave-One-Facet-Out Ranking Robustness}
\label{sec:appendix-lofo}
Main-text strategy rankings use all six facets. Dropping one facet at a time on PDB leaves the SCN Pearson strategy ranking against the external reference unchanged in every drop (all Kendall $\tau=1$ for both $r_f$ and $r_i$). OE profile-correlation ($r_i$) rankings against PDB vary modestly ($\tau=0.8$--$1$), always as adjacent swaps among already-close intermediate strategies; Trait-injected remains first and Raw Dialogue last. Versus the self-report anchor, reported order changes are almost entirely Wiki-based RAG $\leftrightarrow$ Dialogue-driven swaps, which the main text already treats as a near-tie. These checks support ranking robustness \emph{within the selected six-facet space}; they do not establish psychometric reliability or validity of the instrument.

\subsection{Lin's CCC vs.\ Pearson}
\label{sec:appendix-ccc}
Tables~\ref{tab:main-results}--\ref{tab:facets-anchor} report Pearson $r$ / Lin's CCC in each cell. CCC is computed on raw $1$--$5$ scores (not separately standardized) and aggregated in Fisher-$z$ space, matching Pearson. CCC is lower than Pearson on all three datasets, more so for OE than SCN (facet-mean CCC vs.\ $r$: PDB $0.75$/$0.80$ SCN and $0.59$/$0.71$ OE; WildChat $0.72$/$0.79$ and $0.49$/$0.67$; LMSYS $0.61$/$0.69$ and $0.42$/$0.59$), consistent with mean shifts and frequent scale compression of behavioral scores relative to the anchor. Full strategy rankings are metric-sensitive: on LMSYS OE, persona-profile CCC vs.\ Pearson rankings yield Kendall $\tau=0$. The headline pairwise gains remain positive for Trait-injected vs.\ Raw Dialogue on PDB (SCN CCC$_i$ difference $0.20$ $[0.15,0.25]$; OE $0.37$ $[0.31,0.42]$) and USP Profile vs.\ Raw Dialogue on WildChat (SCN $0.05$ $[0.03,0.07]$; OE $0.07$ $[0.04,0.10]$) and LMSYS (SCN $0.07$ $[0.05,0.09]$; OE $0.10$ $[0.07,0.13]$); all six persona-bootstrap $95\%$ CIs exclude $0$. We therefore keep Pearson as the primary ranking statistic and treat CCC as a qualifier of absolute agreement.

\section{MTMM Discriminant Validity: Full Criterion Breakdown}
\label{sec:appendix-mtmm}
This appendix expands the Campbell--Fiske multitrait--multimethod (MTMM) analysis summarized in \mbox{\S\ref{sec:anchor-support}}. On PDB we evaluate the three criteria per facet--method-pair within each of the $20$ (construction-strategy $\times$ answering-model) cells ($6$ facets $\times$ $3$ method pairs $=18$ tests per cell, $360$ tests total; $n=75$ personas per cell; open-ended scores use the default judge, \texttt{DeepSeek-V4-Pro}), and report the fraction of tests that pass each criterion: C1 (validity significantly greater than zero) $99\%$; C2 (validity exceeds heterotrait-heteromethod competitors) $79\%$; C3 (validity exceeds heterotrait-monomethod competitors) $71\%$; all three jointly $68\%$. The mean convergent (monotrait-heteromethod) correlation is $r=0.73$ in Fisher-$z$ space. The failures are facet-specific rather than uniform: across each facet's $60$ tests ($3$ method pairs $\times$ $20$ cells), Respectfulness and Emotional Volatility satisfy all three criteria in every test, whereas Creative Imagination satisfies C2/C3 in only $33\%$/$18\%$ of its tests, consistent with its moderate correlation with the anchor (\mbox{Table~\ref{tab:facets-anchor}}) reflecting shared method variance in addition to trait-specific signal.

\paragraph{Extension to WildChat and LMSYS.} Repeating the identical analysis on WildChat and LMSYS ($4$ strategies $\times$ $4$ models $=16$ cells, $288$ tests each per dataset) yields the same picture: mean convergent $r=0.69$ and $0.61$, with all-three-criteria pass rates of $75\%$ and $63\%$, respectively (the individual C1--C3 breakdown is available only for PDB). Respectfulness and Emotional Volatility again satisfy all three criteria in every one of their tests on both datasets, while Creative Imagination and Intellectual Curiosity remain the weakest facets for discriminant validity (C2/C3 pass rates of roughly $30$--$65\%$). This facet-level pattern therefore recurs across datasets rather than being specific to PDB's well-documented personas.

\section{Persona-Identifiability: Per-Cell Detail and Omnibus Test}
\label{sec:appendix-identifiability}
This appendix expands the persona-identifiability analysis of \mbox{\S\ref{sec:anchor-support}}. For each of the $20$ PDB (construction-strategy $\times$ answering-model) cells (default judge \texttt{DeepSeek-V4-Pro}, $n=75$ personas per cell), we run a one-sided label-permutation test ($10{,}000$ shuffles) on (a) top-1 identification accuracy and (b) the mean percentile rank of the true match. The complete per-cell statistics (accuracy, percentile rank, permutation $z$ and $p$ for both channels) are released with the code; Table~\ref{tab:identif-summary} summarizes their distribution.

\paragraph{Aggregation of permutation-test results.} A mean of permutation $p$-values has no clean sampling interpretation: it is neither a valid combined significance level nor an effect size. We therefore report the \emph{fraction of cells} that individually reject the chance null and the median $[\min,\max]$ per-cell $p$, and complement these with a single omnibus test.

\paragraph{Stratified omnibus permutation test.} To obtain one significance statement per channel while respecting the cell structure, we permute persona labels \emph{within each cell independently and simultaneously}, recompute the per-cell statistic, and aggregate across cells into a single global statistic: the total number of correct top-1 identifications summed over all $20$ cells (equivalently, over all $20\times75=1500$ trials) for the top-1 test, and the trial-pooled mean percentile rank for the rank test. Repeating this $10{,}000$ times yields the null distribution of the global statistic. The observed totals ($\approx231$ correct top-1 identifications for the scenario channel and $\approx154$ for the open-ended channel, versus $\approx20$ expected under the stratified null) lie more than $30$ null standard deviations above chance, so the omnibus test rejects chance at $p<0.001$ for both channels and both statistics. This aggregation accounts for the shared-persona structure across cells (the same $75$ personas recur across strategies and answering models), which a naive pooling of independent tests would ignore.

\begin{table}[t]
\centering\footnotesize
\setlength{\tabcolsep}{2.5pt}
\begin{tabular*}{\columnwidth}{@{\extracolsep{\fill}}l cc cc@{}}
\toprule
& \multicolumn{2}{c}{\textbf{Top-1 acc.}} & \multicolumn{2}{c}{\textbf{Mean pct.\ rank}} \\
\cmidrule(lr){2-3}\cmidrule(lr){4-5}
\textbf{Statistic} & SCN & OE & SCN & OE \\
\midrule
Mean over cells      & $0.15$ & $0.10$ & $0.14$ & $0.20$ \\
Cells sig.\ ($p<0.05$) & $20/20$ & $16/20$ & $20/20$ & $20/20$ \\
Median per-cell $p$  & ${<}0.001$ & ${<}0.001$ & ${<}0.001$ & ${<}0.001$ \\
Max per-cell $p$     & $0.013$ & $0.266$ & ${<}0.001$ & ${<}0.001$ \\
Omnibus $p$          & ${<}0.001$ & ${<}0.001$ & ${<}0.001$ & ${<}0.001$ \\
\bottomrule
\end{tabular*}
\caption{PDB persona-identifiability summary over the $20$ (strategy $\times$ answering-model) cells (default judge \texttt{DeepSeek-V4-Pro}). Chance is $1/75$ for top-1 accuracy and $0.5$ for the mean percentile rank, and lower percentile ranks are better. ``Cells sig.'' is the count of cells whose per-cell label-permutation test rejects chance at $\alpha=0.05$; we report the median and max per-cell $p$ rather than a mean $p$. ``Omnibus $p$'' is the stratified within-cell permutation test described above. The four open-ended top-1 non-significant cells are the weakly-conditioned cells (three \texttt{gpt-oss-120b} cells and Name-only \texttt{HY3-preview}).}
\label{tab:identif-summary}
\end{table}

\paragraph{Extension to WildChat and LMSYS.} Repeating the identical mismatch-control analysis on WildChat and LMSYS shows the same pattern. Scenario-channel mean top-1 accuracy is $9\%$ and $8\%$ (chance $\approx1\%$), with mean percentile ranks of $0.18$ and $0.23$ (percentile-rank test $p<0.001$ in every cell). Open-ended top-1 accuracy is weaker and not significant in every cell, but the open-ended mean percentile rank remains far below chance ($0.27$ and $0.29$, $p<0.001$ in every cell). Persona-level identifiability by percentile rank is therefore not specific to PDB's well-documented personas.

\section{Decoding Settings and Computational Cost}
\label{sec:appendix-compute}

\paragraph{Decoding settings.} Table~\ref{tab:decoding} records the generation parameters used in the reported runs. Requests to all four answering models specified \texttt{top\_p}${}=0.95$ and \texttt{reasoning\_effort}${=}$\texttt{"no\_think"}. SCN and OE answering used temperature $0.2$; the five Likert administrations used $0.15$, $0.20$, $0.25$, $0.15$, and $0.20$. All three judges used temperature $0.1$; judge \texttt{top\_p} and reasoning/thinking parameters were not set in the request and followed serving/API defaults. We distinguish request parameters from serving defaults: only the former are under our control.

\begin{table}[h]
\centering\scriptsize
\setlength{\tabcolsep}{2pt}
\resizebox{\columnwidth}{!}{%
\begin{tabular}{l l l}
\toprule
\textbf{Call type} & \textbf{Temp.} & \textbf{Other request params} \\
\midrule
Likert (5 admins) & $0.15$, $0.20$, $0.25$, $0.15$, $0.20$ & \texttt{top\_p}${}=0.95$, \texttt{no\_think} \\
SCN answering     & $0.2$ & \texttt{top\_p}${}=0.95$, \texttt{no\_think} \\
OE answering      & $0.2$ & \texttt{top\_p}${}=0.95$, \texttt{no\_think} \\
OE judges         & $0.1$ & \texttt{top\_p}/thinking: API default \\
\bottomrule
\end{tabular}%
}
\caption{Decoding settings. Answering-model \texttt{top\_p} and \texttt{reasoning\_effort}${=}$\texttt{"no\_think"} were set in every answering request. Judge \texttt{top\_p} and thinking were left at serving/API defaults.}
\label{tab:decoding}
\end{table}

\paragraph{Models and infrastructure.} All open-weight models are self-deployed on an in-house H20 GPU cluster and served in the precision of their released checkpoints. The four answering models are \texttt{GLM-5.1-FP8} ($744$B parameters, $40$B active per token, served in FP8 on $16$ GPUs), \texttt{Hy3-preview} ($295$B, $21$B active, $8$ GPUs), \texttt{gpt-oss-120b} ($117$B, $5.1$B active, $4$ GPUs) and \texttt{gemma-4-31B-it} ($30.7$B, dense, $2$ GPUs); the two open-weight judges are \texttt{DeepSeek-V4-Pro} ($1.6$T, $49$B active, $16$ GPUs) and \texttt{GLM-5.2} ($744$B, $40$B active, $32$ GPUs). GPU counts are per serving instance. Two externally hosted models are used, whose sizes are not publicly disclosed: \texttt{GPT-5.5} generates the three synthesized persona representations once, offline \mbox{(\S\ref{sec:appendix-construction})}, and \texttt{Gemini-3.1-Pro} re-scores the open-ended responses of the judge-robustness subset \mbox{(\S\ref{sec:model-robust})}. Neither is required to reproduce the main results, which depend only on self-deployed models and on the representation files released with the code.

\paragraph{Calls per persona.} Evaluating one persona under one representation strategy with one answering model and the default judge takes $192$ model calls: $90$ self-report calls ($18$ items $\times$ $5$ administrations), $30$ scenario calls, $18$ open-ended generation calls, and $54$ judging calls ($18$ responses $\times$ $3$ independent judgments). Of these, $138$ fall on the answering model and $54$ on the judge.

\paragraph{Total scale.} The design comprises $4{,}700$ persona $\times$ strategy $\times$ answering-model units ($75 \times 5 \times 4$ on PDB, $100 \times 4 \times 4$ on each of WildChat and LMSYS), giving $648{,}600$ answering calls and $253{,}800$ default-judge calls. The judge-robustness re-scoring covers $3{,}100$ units with two additional judges, adding $334{,}800$ calls. The full experiment therefore issues approximately $1.24$ million model calls, of which $167{,}400$ go to the only closed judge. Representation generation is negligible by comparison, at one \texttt{GPT-5.5} call per conversation and synthesized representation.

\paragraph{Wall-clock and GPU hours.} Timings are measured over $52$ production batches at concurrency $8$ rather than estimated. Per persona-unit, the answering models take $128.4$s \mbox{(\texttt{GLM-5.1-FP8})}, $58.4$s \mbox{(\texttt{Hy3-preview})}, $43.2$s \mbox{(\texttt{gemma-4-31B-it})} and $42.9$s \mbox{(\texttt{gpt-oss-120b})}. Each answering model covers $1{,}175$ units, so the answering side takes about $41.9$, $19.1$, $14.1$ and $14.0$ hours, roughly $89$ wall-clock hours in total; weighting each by the GPUs its instance occupies gives about $907$ GPU hours. The marginal cost of the framework is therefore $0.02$--$0.57$ GPU hours per persona, strategy and answering model, the spread reflecting model size rather than anything specific to the instruments. Judging runs on the same cluster and is reported above as call counts. These figures cover the reported runs and exclude prompt development and discarded configurations.

\end{document}